\documentclass[preprint,12pt]{elsarticle}

\usepackage[english]{babel}
\usepackage[utf8]{inputenc}
\usepackage{times}
\usepackage{epsfig}
\usepackage{graphicx}
\usepackage{amsmath}
\usepackage{amsfonts}
\usepackage{url}
\usepackage{color}
\usepackage{caption}
\usepackage{subcaption}
\usepackage[ruled,vlined]{algorithm2e}
\usepackage[pagebackref=true,breaklinks=true,colorlinks,bookmarks=false]{hyperref}

\DeclareMathOperator*{\argmax}{arg\,max}

\begin{document}

\begin{frontmatter}


\title{Online Adaptive Hidden Markov Model for Multi-Tracker Fusion}


\author[label1]{Tomas Vojir\corref{cor1}}
\ead{vojirtom@cmp.felk.cvut.cz}
\cortext[cor1]{corresponding author}
\ead[url]{http://cmp.felk.cvut.cz/~vojirtom/}

\author[label1]{Jiri Matas}
\ead{matas@cmp.felk.cvut.cz}
\ead[url]{http://cmp.felk.cvut.cz/~matas/}
\address[label1]{The Center for Machine Perception, FEE CTU in Prague\\
Karlovo namesti 13, 121 35 Prague 2, Czech Republic\\}

\author[label2]{Jana Noskova}
\ead{noskova@fsv.cvut.cz}
\address[label2]{Faculty of Civil Engineering, CTU in Prague\\
Thakurova 7/2077, 166 29 Prague 6, Czech Republic\\}


\begin{abstract} 

In this paper, we propose a novel method for visual object tracking
called HMMTxD. The method fuses observations from complementary out-of-the box 
trackers and a detector by utilizing a hidden Markov 
model whose latent states correspond to
a binary vector expressing the failure of individual trackers. The Markov
model is trained in an unsupervised way, relying on an online learned detector
to provide a source of tracker-independent information for a modified Baum-
Welch algorithm that updates the model w.r.t. the partially annotated data.

We show the effectiveness of the proposed method on combination of two and three
tracking algorithms. The performance of HMMTxD is evaluated on two standard
benchmarks (CVPR2013 and VOT) and on a rich collection of 77 publicly
available sequences. The HMMTxD outperforms the state-of-the-art, often
significantly, on all datasets in almost all criteria. 

\end{abstract}


\end{frontmatter}



\section{Introduction}

In the last thirty years, a large number of diverse visual tracking methods
has been proposed~\cite{Yilmaz2006,smeulder2013}. The methods differ in the
formulation of the problem, assumptions made about the observed motion, in
optimization techniques, the features used, in the processing speed, and in
the application domain. Some methods focus on specific challenges like
tracking of articulated or deformable
objects~\cite{Kwon2009,godec2011,Cehovin2013}, occlusion
handling~\cite{grabner2010}, abrupt motion~\cite{Zhou2010} or long-term
tracking~\cite{Pernici2013,Kalal2012}.

Three observations motivate the presented research. First, most trackers
perform poorly if run outside the scenario they were designed for. Second,
some trackers make different and complementary assumptions and their failures
are not highly correlated (called complementary trackers in the paper). And
finally, even fairly complex well performing trackers run at frame rate or
faster on standard  hardware, opening the possibility  for multiple trackers
to run concurrently and yet in or near real-time.

We propose a novel methodology that exploits a hidden Markov model (HMM) for
fusion of non-uniform observables and pose prediction of multiple complementary 
trackers using an on-line learned high-precision detector. The non-uniform 
observables, in this sense, means that each tracker can produce its own 
type of ''confidence estimate'' which may not be directly comparable between each other. 

The HMM, trained in an unsupervised manner,
estimates the state of the trackers -- failed, operates correctly -- and
outputs the pose of the tracked object taking into account the past
performance and observations of the trackers and the detector. The HMM
treats the detector output as correct if it is not in contradiction with its 
current most probable state in which the majority of trackers are correct. This limits the cases 
where the HMM would be wrongly updated by a false detection. For the potentially 
many frames where reliable detector output is not available, it combines the trackers. 
The detector is trained on the first image and interacts with the learning of the
HMM by partially annotating the sequence of HMM states in the time of verified
detections. The recall of the detector is not critical but it affects the
learning rate of the HMM and the long-term properties of the HMMTxD method,
i.e. its ability to reinitialize trackers after occlusions or object
disappearance.

{\bf Related work.} The most closely related approaches include Santner et
al.~\cite{Santner2010}, where three tracking methods with different rates of
appearance adaptation are combined to prevent drift due to incorrect model
updates. The approach uses simple, hard-coded rules for tracker selection.
Kalal et al.~\cite{Kalal2012} combine a tracking-by-detection method with a
short-term tracker that generates so called P-N events to learn new object
appearance. The output is defined either by the detector or the tracker based
on visual similarity to the learned object model. Both these methods employ
pre-defined rules to make decisions about object pose and use one type of
measurement, a certain form of similarity between the object and the estimated
location. In contrary, HMMTxD learns continuously and causally  the  performance statistics
of individual parts of the systems and fuses multiple "confidence"
measurements in the form of probability densities of observables in the HMM.
Zhang et al.~\cite{zhang2014} use a pool of multiple classifiers learned from
different time spans and choose the one that maximize an entropy-based cost
function. This method addresses the problem of model drifting due to wrong
model updates, but the failure modes inherent to the classifier itself remains
the same. This is unlike the proposed method which allows to combine diverse
tracking methods with different inherent failure modes and with different
learning strategies to balance their weaknesses.

Similarly to the proposed method, Wang et al.~\cite{wangg14} and Bailer et
at.~\cite{Bailer2014} fuse different out-of-the box tracking methods. Bailer
et al. combine offline the {\it outputs} of multiple tracking algorithms.
There is no interaction between trackers, which for instance implies that the
method avoids failure only if one method correctly tracks the whole sequence.
Wang et al. use a factorial hidden Markov model and a Bayesian approach. The
state space of their factorial HMM is the set of potential object positions,
therefore it is very large. The model contains a probability description of
the object motion based on a particle filter. Trackers interact by
reinitializing those with low reliability to the pose of the most confident
one. The Yuan et al.~\cite{Yuan2015} using HMM in the same setup, but rather than merging
multiple tracking method, they focus on modeling the temporal change of the target 
appearance in the HMM framework by introducing a observational dependencies. 
In contrast, the HMMTxD method is online with tracker
interaction via a high precision object detector that supervises tracker reinitializations
which happen on the fly. The appearance modeling is performed inside of each tracker and
the HMMTxD capture the relation of the confidence provided by tracker and its performance,
validated by the object detector, by the observable distributions. Moreover, the HMMTxD
confidence estimation is motion-model free and this prevents biases towards support of
trackers with a particular motion model.

Yoon et al.~\cite{Yoon2012} combines multiple trackers in a particle filter
framework. This approach models observables and transition behavior of
individual trackers, but the trackers are self-adapting which makes it prone
to wrong model updates. The adaptation of HMMTxD model is supervised by a detector 
method set to a specific mode of operation -- near $100\%$
precision -- alleviating the incorrect update problem.

The contributions of the paper are: a novel method for fusion of multiple
trackers based on HMMs using non-uniform observables, a simple, and so far unused, unsupervised 
method for HMMs training in the context of tracking, tunable feature-based detector with very 
low false positive rate, and the creation of a tracking system that shows state-of-the-art performance.

\section{Fusing Multiple Trackers}

HMMTxD uses a hidden Markov model (HMM) to integrate pose and observational
confidence of different trackers and a detector, and updates its own
confidence estimates that in turn define the pose that it outputs. In the HMM,
each tracker is modeled as working correctly (1) or incorrectly (0). The HMM
poses no constraints on the definition of tracker correctness, we adopted
target overlap above a threshold.    Having at our disposal $\mathbf{n}$
trackers, the set of all possible states is
$\{s_1,s_2,\ldots,s_N\}=\{0,1\}^{\mathbf{n}}, N=2^{\mathbf{n}}$ and the initial state $s_1=(1,1,\ldots,1)$. 
Note that the trackers are not assumed to be independent, because an independence of tracker
correctness is not a realistic assumption. For example, if the tracking problem
is relatively easy, all trackers tend to be correct and in the case of
occlusion all tend to be incorrect (see the analysis
in~\cite{votpamiarxiv2015}). The number of states $2^{\mathbf{n}}$ grows
exponentially with the number of trackers. However, we do not consider this a
significant issue -- due to ''real-time'' requirements of tracking, the need
to combine more than a small number of trackers, say ${\mathbf{n}}=4$, is
unlikely.

\begin{figure}
    \centering
    \includegraphics[width=0.85\textwidth]{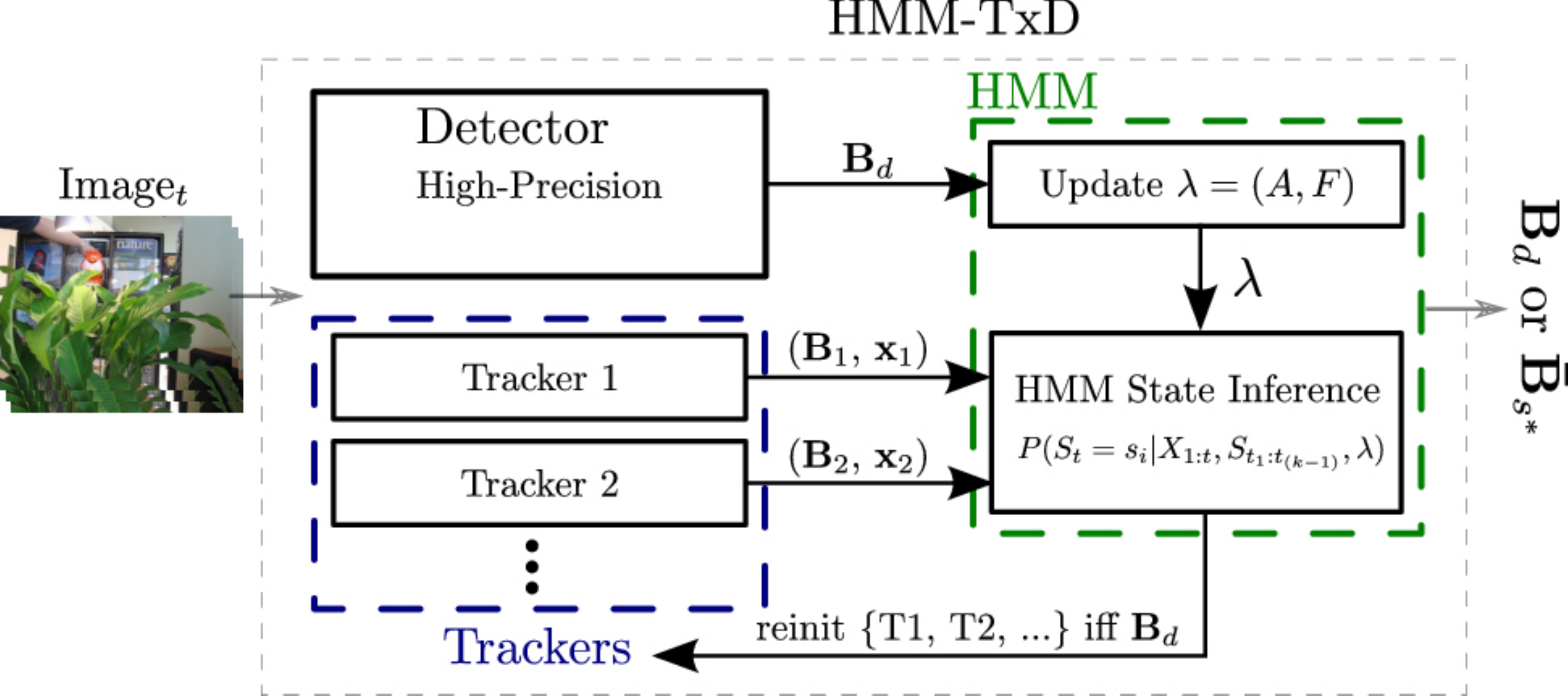}
\caption{
The structure of the HMMTxD. For each frame, the detector and trackers are
run. Each tracker outputs a new object pose and observables $(\mathbf{B}_i,
\mathbf{x}_i)$ and the detector outputs either the verified object pose
$\mathbf{B}_d$ or nothing. If detector fires, HMM is updated and trackers are
reinitialized and the final output is the $\mathbf{B}_d$, otherwise, HMM
estimate the most probable state $s^*$ and outputs an average bounding box
$\mathbf{\bar{B}}_{s^*}$ of trackers that are correct in the estimated state
$s^*$.}

\label{fig:method}
\end{figure}

The HMMTxD method overview is illustrated in Fig.~\ref{fig:method}. Each
tracker provides an estimate of the object pose ($\mathbf{B}_i$) and a vector
of observables ($\mathbf{x}_i$), which may contain a similarity measure to
some model (such as normalized cross-correlation to the initial image patch,
distance of template and current histograms at given position, etc.) or any
other estimates of the tracker performance. The $\mathbf{x}_i,
i=\{1,2,\ldots,\mathbf{n}\}$ serve as observables to relate the tracker current 
confidence to the HMM. Each individual
observable depends only on one particular tracker and its correctness, hence,
they are assumed to be conditionally independent conditioned on the state of
the HMM (which encodes the tracker correctness).

In general, there are no constraints on observable values, however, in the proposed HMM the observable values are required to be normalized to the $(0,1)$ interval. The observables 
are modeled as beta-distributed random variables (Eq.~\ref{eq:beta}) and its parameters
are estimated online.  The beta
distribution was chosen for its versatility, where practically any kind of
unimodal random variable on $(0,1)$ can be modeled by the beta distribution,
i.e. for any choice of any lower and upper quantiles, a beta distribution
exists satisfying the given quantile constraint~\cite{gupta2004}. 

Learning the parameters of the beta distributions online is crucial 
for the adaptability to particular tracking scenes, where the observable values from
a different trackers may be biased due to scene properties, or to adapt to a different types of observables of trackers and their correlations to the ''real'' tracker performance. For example, taking correlation with the initial target patch as an observable for one tracker and color histogram distance to a initial target for a second tracker, the correlation between their values and the performance of the tracker may differ depending on object rigidity and color distribution of object and background.

The HMM is parameterized by the pair $\lambda = (A, F)$, where $A$ are the probabilities
of state transition and $F$ are the beta distributions of observables with shape parameters $p,q>0$ and density defined for $x \in (0,1)$

\begin{equation} 
f(x|p,q) = \frac{x^{p-1}(1-x)^{q-1}}{\int_{0}^{1} u^{p-1}(1-u)^{q-1}du}.
\label{eq:beta}
\end{equation}

Since the goal is real-time tracking without any specific pre-processing,
learning of HMM parameters has to be done online. Towards this goal, the object detector, which
is set to operating mode with low false positive rate, is utilized to
partially annotate the sequence of hidden states. In contrast to classical
HMM, where only a sequence of observations $\mathbb X=\{X_t\}_{t=1}^{T},X_t=(\mathbf{x}_1,\mathbf{x}_2,\ldots,\mathbf{x}_\mathbf{n})$ is available, we are
in a semi-supervised setting and have a time sequence $0=t_0<t_1<t_2\ldots
<t_K\leq T$ of observed states of a Markov chain
$\mathbb{S}=\{S_{t_{k}}=s_{i_k},\{t_k\}_{k=1}^{K}\}$, and Markov chain
starting again in state $s_1$, all trackers correct, at any time $\{t_{k}+1,0\leq k\leq K\}$, since there are reinitialized to common object pose. This
information is provided by the detector, where $\{t_k\}_{k=1}^{K}$ is a
sequence of detection times. The HMM parameters are learned by a modified
Baum-Welch algorithm run on the observations $\mathbb X$ and the annotated
sequence of states $\mathbb{S}$. The partial annotation and HMM parameter
estimation update is done strictly online.

\begin{figure}[!t]
    \centering
    \includegraphics[width=0.85\textwidth]{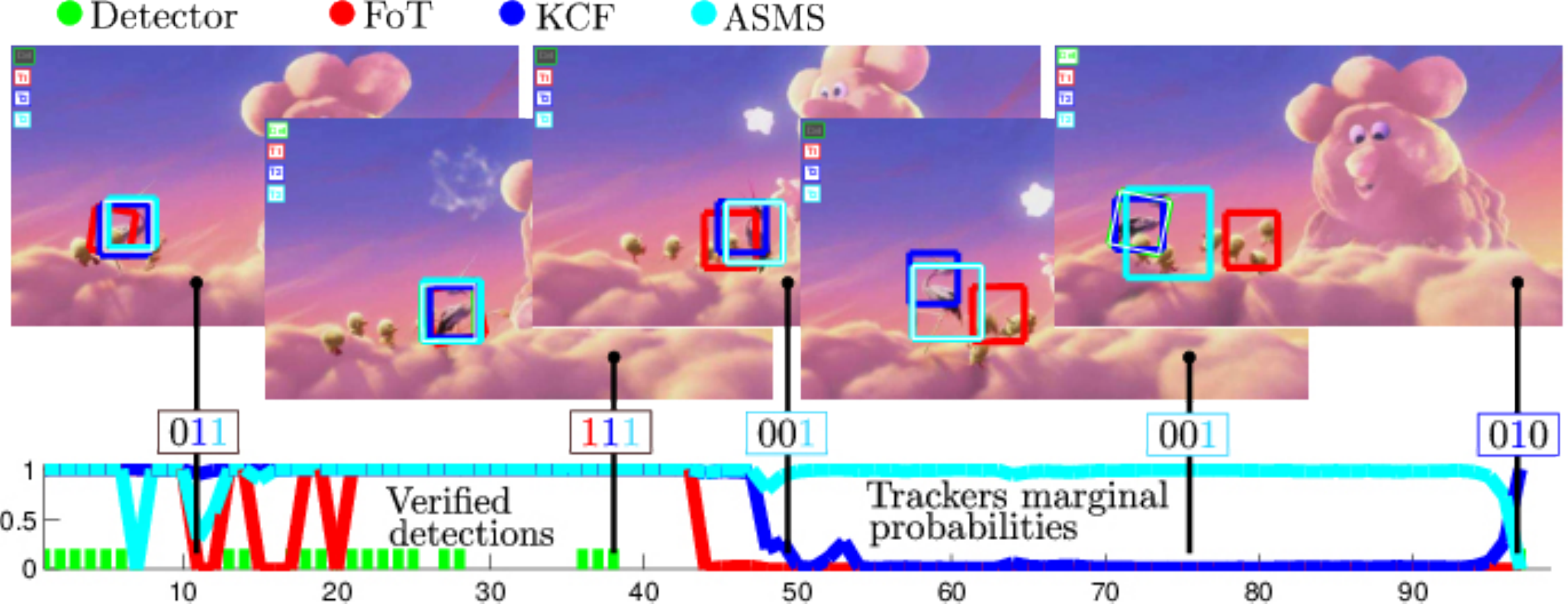}
\caption{
Illustration of HMM state and trackers probability estimation during tracking.
The bottom graph shows the marginal probabilities for each tracker being
correct and the detection times (green spikes). Above the graph the inferred
states $s^*_t$ with color encoded correct trackers (1) are displayed. The
final output is defined by the state  $s^*_t$ and the bounding box is
highlighted by white color. Best viewed zoomed in color.}

\label{fig:insight}
\end{figure}

The output of the HMMTxD is an average bounding box of correct trackers of the
current most probable state $s_t^*$. For $t_{(k-1)}<t< t_k, 1 \le k \le K$ the
forward-backward procedure~\cite{Rabiner1989} for HMM is used to calculate
probability of each state at time $t$ (see Eq. \ref{eq:A1}-\ref{eq:A7}) and the state $s_t^* \in
\{0,1\}^{\mathbf{n}}\setminus (0,0,\ldots,0)$ is the state for which

\begin{equation} 
P(S_{t}=s_i|X_{1},\ldots, X_{t},S_{t_1},\ldots, S_{t_{(k-1)}},\lambda)
\label{eq:state-prob}
\end{equation}

is maximal. This equation is computed using Eq.~\ref{eq:app_chain} and maximized w.r.t $i, 1 \leq i \leq N$. For $t_{K}<t \leq T$ the Eq.~\ref{eq:state-prob} holds with
$t_{(k-1)}=t_K$. This ensures that the algorithm outputs a pose for each frame
which is required by most benchmark protocols. Illustration of the tracking
process and HMM insight is shown in Fig.~\ref{fig:insight}. Theoretically the
parameters of HMM could be updated after each frame. However, in our
implementation, learning takes place only at frames where the detector
positively detects the object, i.e. the sequence of states starting and ending with observed state inferred by the detector\footnote{If pure online fusion
is not required, future observations can also be used to determine the probability
of each state.}.
The detector is used only if the detection pose is not in contradiction with the pose of the current most probable state in which the majority of trackers are correct. This ensure that even when the detector makes a mistake, the HMM is not wrongly updated. When we are in the state that one or none of the trackers are correct, the detector get precedence.

\section{Learning the Hidden Markov Model}
\label{sec:hmm}

For learning of the parameters $\lambda$ of the HMM a MLE inference is
employed, however maximizing the likelihood function
$P(\mathbb{X},\mathbb{S}|\lambda)$ is a complicated task that cannot be solved
analytically. In the proposed method, the Baum-Welch algorithm~\cite{baum1970}
is adapted. The Baum-Welch algorithm is a widespread iterative procedure for
estimating parameters of HMM where each iteration increases the likelihood
function but, in general, the convergence to the global maximum is not
guaranteed. The Baum-Welch algorithm is in fact an application of the EM
(Expectation-Maximization) algorithm~\cite{Dempster77}.

\subsection{Classical Baum-Welch Algorithm}
Let us assume the HMM with $N$ possible states
$\{s_1,s_2,\ldots,s_N\}$, the matrix of state transition probabilities
$A=\{a_{ij}\}_{i,j=1}^{N}$, the vector of initial state probabilities
$\pi=(1,0,0,\ldots,0)$, the initial state $s_1=(1,1,\ldots,1)$, a sequence of
observations $\mathbb{X}=\{X_t\}_{t=1}^{T},X_t\in R^{m}$ and
$F=\{f_i(x)\}_{i=1}^{N}$ the system of conditional probability densities of
observations conditioned on $S_t=s_i$

\begin{equation}
f_i(x)=f(x |S_t=s_i) \text{ for } 1\leq i\leq N, 1\leq t\leq T, x\in R^m
\end{equation}

where $S_t$ are random variables representing the state at time $t$, and
$\lambda=(A,F)$ is denoting the parameter set of the model.

Let us denote
\begin{equation}
Q(\lambda,\lambda')=\sum_{\mathfrak{s}\in \mathfrak{S}} P(\mathfrak{s}|\mathbb{X},\lambda) \log [P(\mathfrak{s},\mathbb{X}|\lambda')],
\label{eg:q_baum}
\end{equation}
where ${\mathfrak{S}}=\{s_1,s_2,\ldots,s_N\}^T$ is a set of all possible T-tuples of states and \\
$\mathfrak{s}\in \mathfrak{S},\mathfrak{s}=(\mathfrak{s}_1,\ldots,\mathfrak{s}_t,\ldots,\mathfrak{s}_T)$ is one sequence of states.
According to Theorem 2.1. in ~\cite{baum1970}
\begin{equation}
Q(\lambda,\lambda')\geq Q(\lambda,\lambda) \Rightarrow P(\mathbb{X}|\lambda')\geq P(\mathbb{X}|\lambda)
\end{equation}
and the equality holds if and only if
$P(\mathfrak{s}|\mathbb{X},\lambda)=P(\mathfrak{s}| \mathbb{X},\lambda') \text{ for } \forall \mathfrak{s}\in \mathfrak{S}$.
The classical Baum-Welch algorithm repeats the following steps until convergence:
\begin{enumerate}
\item Compute $\lambda^* = \argmax_{\lambda}Q(\lambda_n,\lambda)$
\item Set $\lambda_{n+1}=\lambda^*$.
\end{enumerate}

\subsection{Modified Baum-Welch Algorithm}
We propose the modified Baum-Welch algorithm that exploits the partially annotated sequence of states, where the known states are inferred from the detector output. Let
$0=t_0<t_1<t_2\ldots <t_K\leq T$ be a sequence of detection times,
$\mathbb{S}=\{S_{t_{k}}=s_{i_k},\{t_k\}_{k=1}^{K}\}$ be observed states of
Markov chain, marked by the detector, and $S_{t_{k}+1}=s_1$ for $0\leq k\leq
K$. So the sequence of observations of the HMM is divided into $K+1$
independent subsequences, each with a fixed initial state $s_1$, the first $K$
subsequences with a known terminal state defined by the detector and the last
subsequence with an unknown terminal state.

The following equations are obtained by employing the modification to the Baum-Welch algorithm, 
\begin{equation}
\log [P(\mathfrak{s},\mathbb{X},\mathbb{S}|\lambda)]=\sum_{t=1}^{T-1}\log a_{\mathfrak{s}_{t}\mathfrak{s}_{t+1}} +\sum_{t=1}^{T} \log f_{\mathfrak{s}_{t}}(X_{t}),
\end{equation}
\begin{equation}
\begin{split}
Q(\lambda_n,\lambda)=\sum_{\mathfrak{s}\in \mathfrak{S}} P(\mathfrak{s}|\mathbb{X},\mathbb{S},\lambda_n) \sum_{t=1}^{T-1}\log a_{\mathfrak{s}_{t}\mathfrak{s}_{t+1}} + \\
\sum_{\mathfrak{s}\in \mathfrak{S}} P(\mathfrak{s}|\mathbb{X},\mathbb{S},\lambda_n) \sum_{t=1}^{T} \log f_{\mathfrak{s}_{t}}(X_{t}).
\end{split}
\label{eq:Q}
\end{equation}

The maximization of the $Q(\lambda_n,\lambda)$ can be separated to
maximization w.r.t. transition probability matrix $A=\{a_{ij}\}_{i,j=1}^{N}$ by maximizing 
the first term and w.r.t. observable densities $F=\{f_i(x)\}_{i=1}^{N}$ by maximizing the 
second term.

The maximization of Eq.~\ref{eq:Q} w.r.t. $A$ constrained by $\sum_{j=1}^{N}a_{ij}=1  \text{ for } 1\leq i\leq N$ is obtained by re-estimating the parameters
$\hat{A}=\{\hat{a}_{ij}\}_{i,j=1}^{N}$ as follows:

\begin{equation*}
\hat{a}_{ij}=\frac{\text{expected number of transitions from state $s_i$ to state $s_j$}}{\text{expected number of transitions from state $s_i$}}
\end{equation*}
\begin{equation}
=\frac{\sum_{(t=1\text{ and } t\neq t_{k},1\leq k \leq K) }^{T-1}P(S_{t}=s_i,S_{t+1}=s_j|\mathbb{X},\mathbb{S},\lambda)}
{\sum_{(t=1\text{ and } t\neq t_{k},1\leq k \leq K)}^{T-1}P(S_{t}=s_i|\mathbb{X},\mathbb{S},\lambda)}.
\label{eq:baumWelch}
\end{equation}

This equation is computed using modified forward and backward variables of the
Baum-Welch algorithm to reflect the partially annotated states. For the exact
derivation of formulas for computation of $\hat{a}_{ij}$ see the~\ref{app:fb}.

\subsubsection{Learning Observable Distributions}
The maximization of Eq.~\ref{eq:Q} w.r.t. $F=\{f_i(x)\}_{i=1}^{N}$ depends on
assumptions on the system of probability densities $F$. It is usually assumed
(e.g. in~\cite{Rabiner1989,baum1970}) that $F$ is a system of probability
distributions of the same type and differ only in their parameters.  

In the HMMTxD the $m$-dimensional observed random variables \\
$X_t=(X_t^1,X_t^2,\ldots, X_t^m) \in R^m$ are assumed conditionally
independent and to have the beta-distribution, so $f_i(x) ,1\leq i \leq N$
are products of $m$ one-dimensional beta distributions with parameters of
shape $\{({p_i}^j,{q_i}^j)\}_{j=1}^{m},1\leq i \leq N$. In this case
maximization of the second term of the Eq.~\ref{eq:Q} is an iterative
procedure using inverse digamma function which is very computationally
expensive~\cite{gupta2004}.

We propose to estimate the shape parameters of the beta distributions with a generalized
method of moments. The classical method of moments is based on the fact that
sample moments of independent observations converge to its theoretical ones
due to the law of large numbers for independent random variables. In the
HMMTxD observations $\mathbb{X}=\{X_t\}_{t=1}^{T}$ are not independent. The
generalized method of moments is based on the fact that
$\{X_t-E(X_t|X_1,X_2,\ldots,X_{t-1})\}_{t=1}^{T}$ is a sequence of martingale
differences for which the law of large numbers also holds. Using the generalized
method of moments gives estimates of the parameters of shape

\begin{equation}
\hat{p}_i^j={\hat{\mu}_i^j}\left( \frac{\hat{\mu}_i^j(1-\hat{\mu}_i^j)}{(\hat{\sigma}_i^j)^2}-1\right)
\label{eq:Fp}
\end{equation}
and
\begin{equation}
\hat{q}_i^j=(1-\hat{\mu}_i^j) \left( \frac{\hat{\mu}_i^j(1-\hat{\mu}_i^j)}{(\hat{\sigma}_i^j)^2}-1\right)
\label{eq:Fq}
\end{equation}
where
\begin{equation}
\hat{\mu}_i^j=\frac{\sum_{t=1}^{T}X_t^j P(S_{t}=s_i|\mathbb{X},\mathbb{S},\lambda)}
{\sum_{t=1}^{T}P(S_{t}=s_i|\mathbb{X},\mathbb{S},\lambda)}
\end{equation}
and
\begin{equation}
(\hat{\sigma}_i^j)^2=\frac{\sum_{t=1}^{T}
(X_t^j-\hat{\mu}_i^j)^2P(S_{t}=s_i|\mathbb{X},\mathbb{S},\lambda)}
{\sum_{t=1}^{T}P(S_{t}=s_i|\mathbb{X},\mathbb{S},\lambda)}.
\end{equation}

Let us denote the system of probability densities with re-estimated parameters as
$\hat{F}=\{\hat{f}_i(x)\}_{i=1}^{N}$.
The generalized method of moments is described in detail in the~\ref{app:moments}.


\subsubsection{Algorithm Overview}
The complete modified Baum-Welch algorithm is summarized in
Alg.~\ref{alg:mbw}, where after each iteration
$P(\mathbb{X},\mathbb{S}|\lambda_{n+1})\geq
P(\mathbb{X},\mathbb{S}|\lambda_n)$ and we repeat these steps until
convergence. Note that $\hat{A}_{n}$ is a maximum likelihood estimate of $A$
therefore always increases $P(\mathbb{X},\mathbb{S}|\lambda_n)$ (shown
in~\cite{Rabiner1989}) but $\hat{F}_{n}$ is estimated by the method of moments
so the test on likelihood increase is required (''if statement'' in the
Alg.~\ref{alg:mbw}). In fact, this algorithm structure match to the
generalized EM algorithm (GEM) introduced in ~\cite{Dempster77}.

\begin{algorithm}[h]
\DontPrintSemicolon
 \KwIn{$\mathbb{X},\mathbb{S},\lambda_n=(A_n,F_n)$}
 \KwOut{$\lambda_{n+1}=(A_{n+1},F_{n+1})$}
 \Repeat{convergence $\; \vee \;$ max number of iteration}
 {
    Compute likelihood $P(\mathbb{X},\mathbb{S}|\lambda_n)$\;
    Estimate $\hat{A}_{n}$ by Eq.~\ref{eq:baumWelch} and $\hat{F}_{n}$ by Eq.~\ref{eq:Fp},~\ref{eq:Fq}\;
    \eIf {$P(\mathbb{X},\mathbb{S}|\hat{A}_{n},\hat{F}_{n})<P(\mathbb{X},\mathbb{S}|{A}_{n},{F}_{n})$}
	{
    	$\lambda_{n+1}=(\hat{A}_{n},F_{n})$
	}{
   		$\lambda_{n+1}=(\hat{A}_{n},\hat{F}_{n})$
    }
    $\lambda_{n}=\lambda_{n+1}=(A_{n+1}, F_{n+1})$\;
 }
 \caption{Algorithm for HMM parameters learning}
 \label{alg:mbw}
\end{algorithm}

\section{Feature-Based Detector}
\label{sec:det}

The requirements for the detector are: adjustable operation mode (e.g. set for
high precision but possibly low recall), (near) real-time performance and the
ability to model pose transformations up to at least similarity (translation,
rotation, isotropic scaling). Basically, any detector-like approach can be
used and it may vary based on application. We choose to adapt a feature-based
detector which has been shown to perform well in image retrieval, object
detection and object tracking~\cite{Pernici2013} tasks.

There are many possible combinations of features and their descriptors with
different advantages and drawbacks. We exploit multiple feature types:
specifically, Hessian keypoints with the SIFT~\cite{Lowe2004} descriptor,
ORB~\cite{Rublee2011} with BRISK and ORB with FREAK~\cite{Ortiz2012}. Each
feature type is handled separately, up to the point where point
correspondences are established. A weight is assigned to each feature type
$w^g$ and is set to be inversely proportional to the number of features on the
reference template, to balance the disparity in individual feature numbers.

The detector works as follows. In the initialization step, features are
extracted from the inside and the outside of the region specifying the tracked
object. Descriptors of the features outside of the region are stored as the
background model. 

Usually, the input region is not $100\%$ occupied by the
target; therefore, fast color segmentation~\cite{Kristan2014a} attempts to delineate 
the object more precisely than the axis-aligned bounding box to
remove the features that are most likely not on the target. The step is not critical for the function of the detector, since the bounding box is a fall-back option. We assume that at least $50\%$ of the bounding box is filled with pixels that
belong to the target, if the segmentation fails (returns a region containing less than
$50\%$ of area of the bounding box), all features in the initial bounding box are used.

Additionally, for each target feature, we use a normal distribution $\mathcal{N}(\mu^{f}, \sigma^{f})$ to model the similarity of the feature to other features. The parameters $\mu^{f}$
and $\sigma^{f}$ are estimated in the first frame by randomly sampling 100 features, other 
than $f$, and computing distances to the feature $f$, from which the mean and variation are
computed. This allows defining the quality of
correspondence matches in a probabilistic manner for each feature, thus
getting rid of global static threshold for the acceptable correspondence
distance.

In the detection phase, features are detected and described in the whole
image. For each feature $g_i$ from the image the nearest neighbour (in
Euclidean space or in Hamming distance metric space, depending on the feature
type) feature $b^*$ from the background model and the nearest neighbour
feature $f^*$ from the foreground model are computed. A tentative
correspondence is formed if  the feature match passes the second nearest
neighbour test and a probability that the correspondence distance belongs to
the outlier distribution is lower than a predefined significance set to
$0.1\%$. So

\begin{equation}
\frac{d(g_i, f^*)}{d(g_i, b^*)} < 0.8 \; \wedge \;\mathcal{F} (d(g_i, f^*)| \mu^{f^*}, \sigma^{f^*}) < 0.1\%
\end{equation}
where $\mathcal {F}(d|\mu^{f^*} ,\sigma^{f^*})$ is a c.d.f. of the normal distribution with parameters $\mu^{f^*}$ and 
$\sigma^{f^*}$ of a distance distribution of features not corresponding to $f^*$. 
The $0.1\%$ significance corresponds to the $\mu - 3\sigma$ threshold.
Finally, RANSAC estimates the target current pose using a sum of weighted inliers as a cost function for model support
\begin{equation}
\text{cost} = \sum_i{w^{g_i}*[g_i == \mbox{inlier}] },
\end{equation}
which takes into account the different numbers of features per feature type on the target.

The decision whether the detected pose is considered correct depends on the
number of weighted inliers that supports the RANSAC-selected transformation
and it controls the trade-of between precision and recall of the method. 
This threshold is automatically computed in the first frame of the sequence as $\max(5, \min(0.03*\text{max\_number\_of\_features\_in\_target\_bbox}, 10))$.
The threshold interval (5,10) and the feature multiplier (0.03) were set experimentally to have the false positive rate close to zero for the most of the testing sequences. Furthermore,
majority voting is used to verify that the detection is not in contradiction
to the estimated HMM state, i.e. if we are in the state where two or more (majority)
trackers are correct and the detector is not consistent with them, the
detection is not used. This mitigates the false positive detections, therefore HMM updates, when the trackers works correctly.

\begin{figure*}
    \centering
    \includegraphics[width=\textwidth]{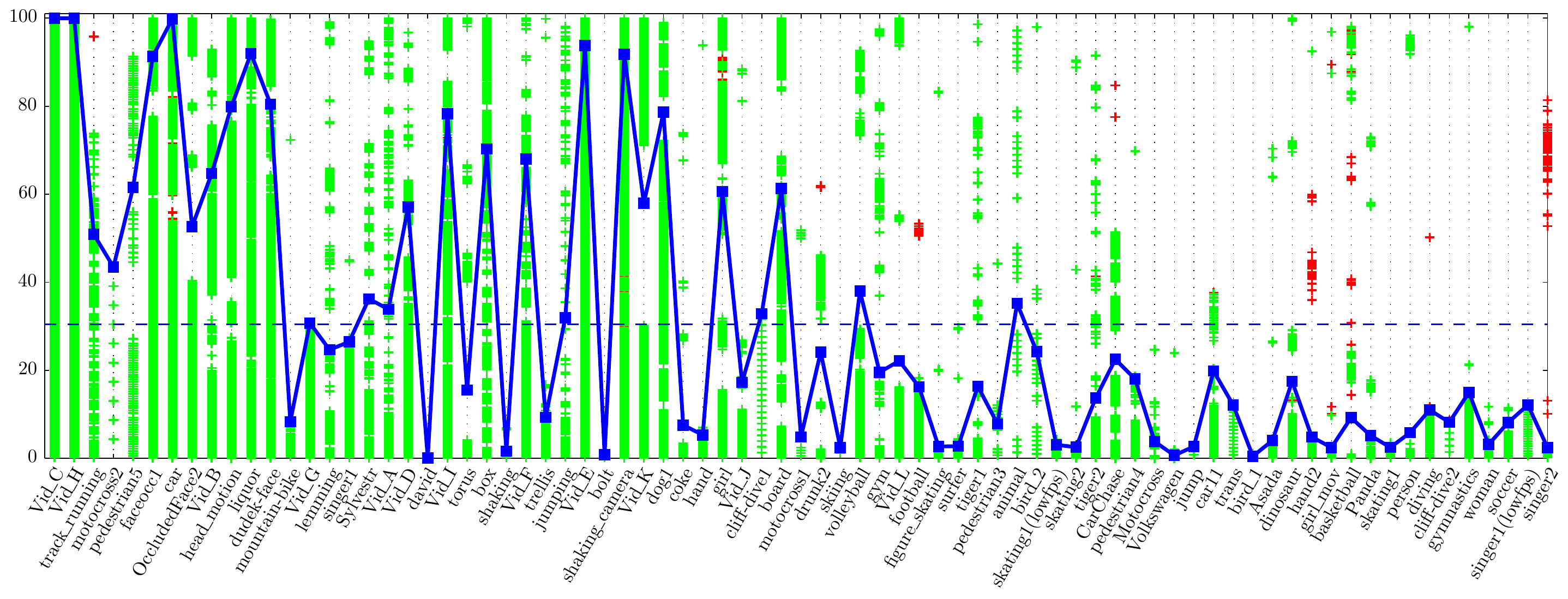}
\caption{
Frames with the detections for 77 sequences dataset. The green marks show the
true positive detection and red marks are false positive. The blue line shows
the recall of the detector and blue dashed line shows the average recall over
all sequences. The length of each sequence is normalized to range $(0,100)$.}

\label{fig:det_perf}
\end{figure*}

The true and false positives for 77 sequences are shown
in Fig.~\ref{fig:det_perf}, where the detector works on almost all sequences
with zero false positive rate ($0.46\%$ average false positive rate on the
dataset) and $30\%$ recall rate.  The failure cases of this feature-based
detector are mostly caused by the imprecise initial bounding box, which
contains large portion of structured background (i.e. background where the
detector finds features) and due to the presence of similar object in the
scene, e.g. sequences \textit{hand2, basketball, singer2}.

\section{HMMTxD Implementation}

To demonstrate the performance of the proposed framework, a pair and a triplet
of published short-term trackers were plugged into the framework to show the
performance gain by combination of a different number of trackers.  As Bailer
et al.~\cite{Bailer2014} pointed out, not all trackers when combined can
improve the overall performance (i.e. adding tracking method with similar
failure mode will not benefit).

We therefore choose methods that have a different designs and work with
different assumptions (e.g. rigid global motion vs. color mean-shift
estimation vs. maximum correlation response). These trackers are the Flock of
Trackers (FoT)~\cite{Vojir2014}, scale adaptive mean-shift tracker
(ASMS)~\cite{Vojir2013} and kernelized correlation filters
(KCF)~\cite{henriques2015}. This choice shows that superior performance can be
achieved by using simple, fast trackers (above 100fps) that may not represent
the state-of-the-art. The trackers can be arbitrarily replaced depending on
the user application or requirements.

\subsection*{Trackers}

The Flock of Trackers (FoT)~\cite{Vojir2014} evenly covers the object with
patches and establishes frame-to-frame correspondence by the Lucas-Kanade
method~\cite{Lucas1981}. The global motion of the target is estimated by
RANSAC.

The second tracker is a scale adaptive mean-shift tracker
(ASMS)~\cite{Vojir2013} where the object pose is estimated by minimizing the
distance between RGB histograms of the reference and the candidate bounding
box. The KCF~\cite{henriques2015} tracker learns a correlation filter by ridge
regression to have high response to target object and low response on
background. The correlation is done in the Fourier domain which is very
efficient.

These three trackers have been selected since they are complementary by
design. FoT enforces a global motion constrain and works  best for rigid
object with texture. On the other hand, ASMS does not enforce object rigidity
and is well suited for articulated or deformable objects assuming their color
distribution is discriminative w.r.t. the background. KCF can be viewed as a 
tracking-by detection approach using sliding window like scanning.

For each tracker position, two global observable measurements are computed,
namely the Hellinger distance between the target template histogram and the
histogram of the current position and normalized cross-correlation score of
the current patch and the target model patch. These target models are
initialized in the first frame and then updated exponentially with factor of
$0.5$ during each positive detection of the detector part. Additionally, each
tracker produces its own estimate of performance. For FoT it is the number of
predicted correspondences (for details please see~\cite{Vojir2014}) that
support the global model. For ASMS it is the Hellinger distance between its
histogram model and current neighbourhood background (i.e. color similarity of
the object and background) and for KCF it is a correlation response of the
tracking procedure.

\begin{figure}[!ht]
    \centering
	\begin{subfigure}[b]{0.49\textwidth}
	    \includegraphics[width=\textwidth]{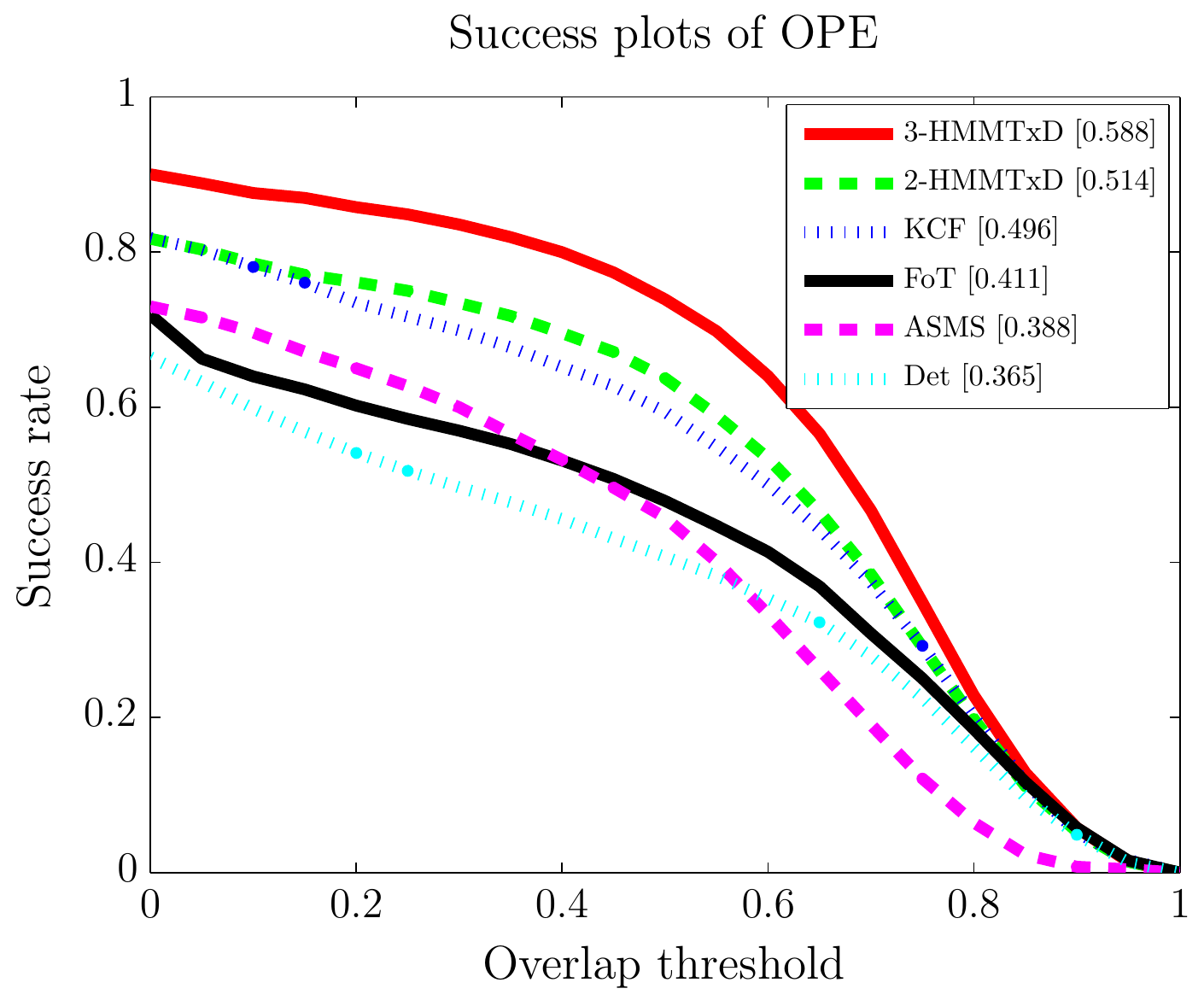}
    \end{subfigure}
    \begin{subfigure}[b]{0.49\textwidth}
    		\includegraphics[width=\textwidth]{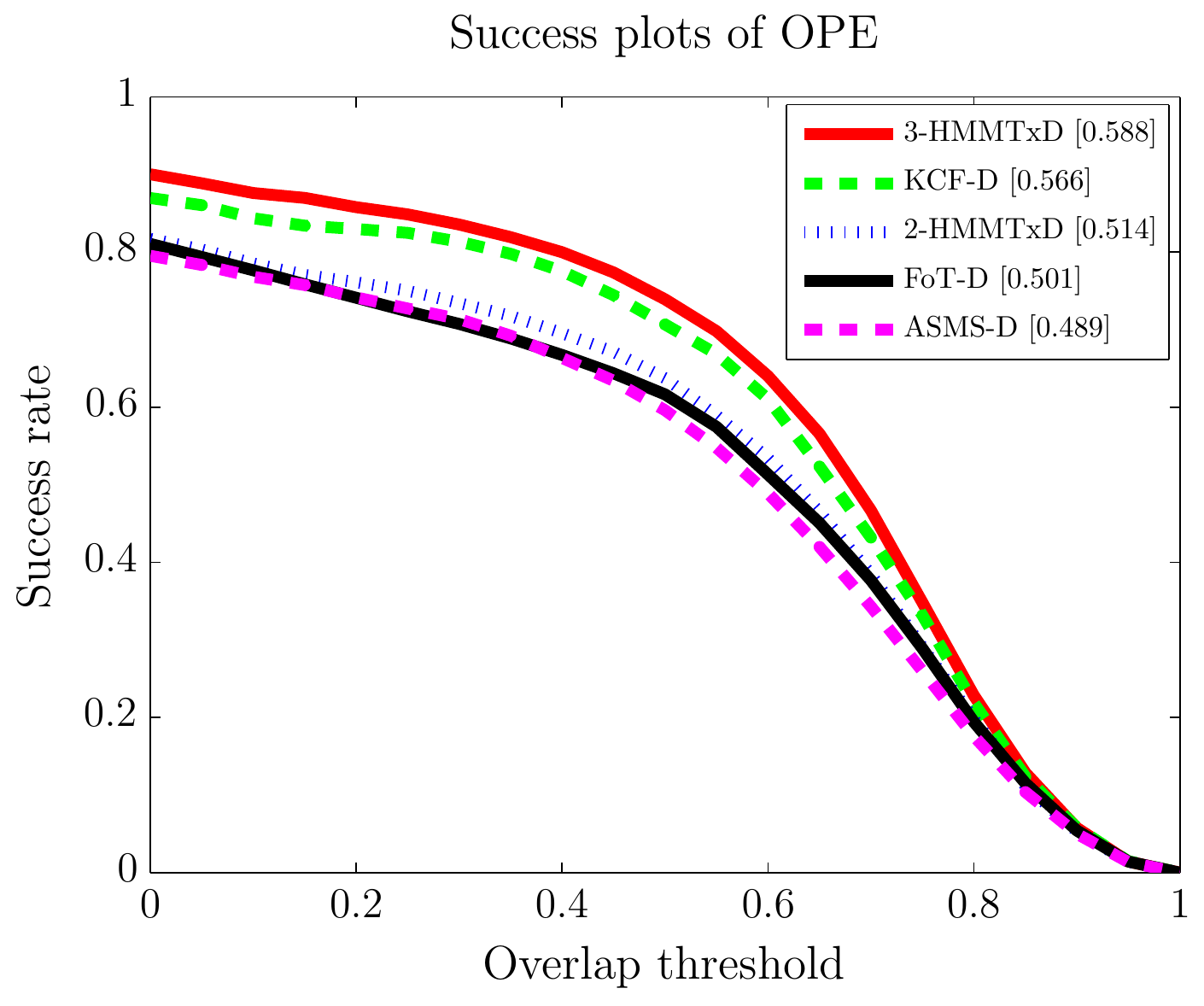}
    \end{subfigure}
\caption{
CVPR2013 OPE benchmark comparison of individual trackers and their combination
in the proposed HMMTxD. The 2-HMMTxD denotes the combination of FoT and ASMS
trackers and 3-HMMTxD is a combination of FoT, ASMS and KCF trackers. Det
stands for the proposed detector. The right plot show simple combination of individual trackers with the proposed detector. Suffix ''-D'' refers to the combination with detector.}

\label{fig:Bench_parts}
\end{figure}

\section{Experiments}

The HMMTxD was compared with state-of-the-art methods on two standard
benchmarks and on a dataset
TV77\footnote{\url{http://cmp.felk.cvut.cz/~vojirtom/dataset/index.html}}
containing 77 public video sequences collected from tracking-related
publications. The dataset exhibits wider diversity of content and variability
of conditions than the benchmarks. 

Parameters of the method were fixed for all
the experiments. In the HMM, the initial beta distribution shape parameters
$(p, q)$ were set to $(2,1)$ for correct state (1) and $(1,2)$ for fail state
(0) for all observations and the transition matrix was set to prefer staying
in the current state. The transition matrix has $0.98$ on diagonal, $0$ in fist column, $0.001$ in last column, $1e-10$ in last row and $0.05$ otherwise. The 
matrix is normalized so that rows sum to one. States in the matrix are binary encoded starting from the left column which corresponds to the state $s_1 = (1, ..., 1)$. The number of iteration for Baum-Welch alg. was set to~$3$.

The processing speed on the VOT2015 dataset is (in frames per second) minimum 1.03, maximum
33.72 and average 10.83 measured on a standard notebook with Intel Core-i7 processor. This
speed is mostly affected by the number of features detected in the images which correlates to
the resolution of the image (in the dataset the range is from 320x180 to 1280x720).

\begin{figure*}[!ht]
    \centering
    \includegraphics[width=\textwidth]{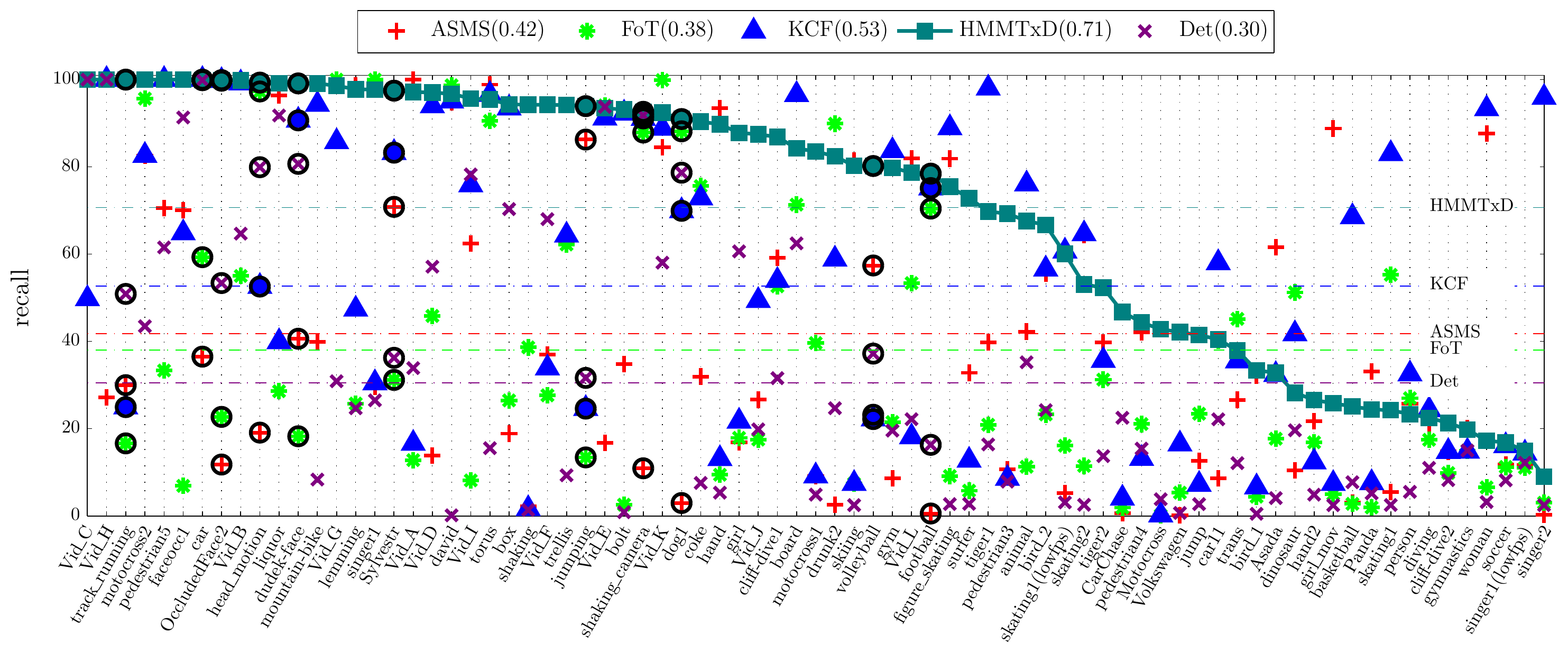}
\caption{
Per sequence analysis of the single trackers (i.e. KCF, ASMS, FoT) and the proposed HMMTxD. The average recall is shown by the dashed lines (precise number is in the legend). Black circles mark grayscale sequences. The sequences are ordered by HMMTxD performance.
}
\label{fig:hmm_parts_single}
\end{figure*}

\begin{figure*}[!ht]
    \centering
    \includegraphics[width=\textwidth]{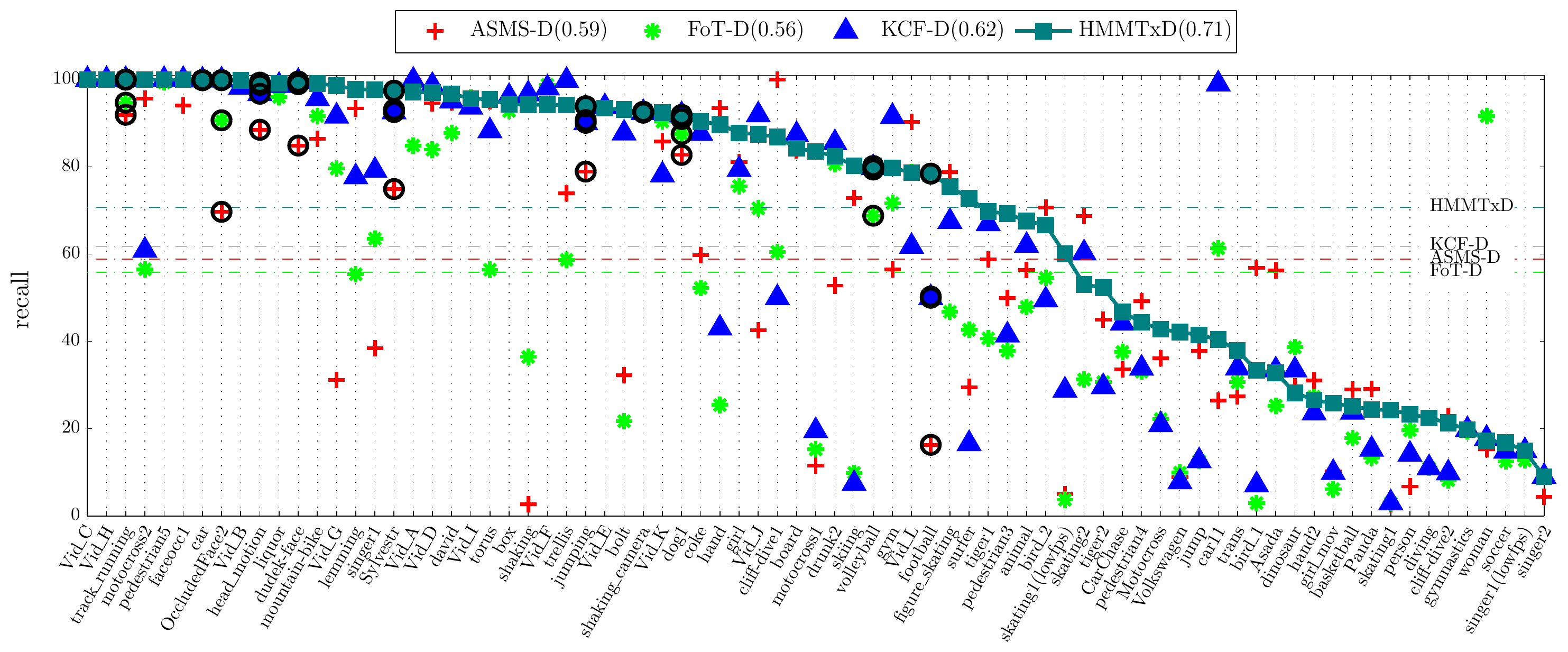}
\caption{
Per sequence analysis of the single trackers combined with the detector (i.e. KCF-D, ASMS-D, FoT-D) and the proposed HMMTxD. The average recall is shown by the dashed lines (precise number is in the legend). Black circles mark grayscale sequences. The sequences are ordered by HMMTxD performance.
}
\label{fig:hmm_parts_det}
\end{figure*}

First, we compare the performance of individual parts of the HMMTxD framework
(i.e. KCF, ASMS, FoT trackers) and their combination via HMM as proposed in
this paper. Two variants of HMMTxD are evaluated -- 2-HMMTxD refers to
combination of FoT and ASMS trackers and the 3-HMMTxD to combination of all
mentioned trackers. We also show the benefit of the proposed detector when simply combined with 
the individual trackers in such way that if detector fires the tracker is reinitialized. 
The Figure~\ref{fig:Bench_parts} shows the benefit gained from the detector and 
further consistent improvement achieved by the combination of the trackers. More detailed per sequence analysis
on the TV77 dataset~(Fig.~\ref{fig:hmm_parts_single} and Fig.~\ref{fig:hmm_parts_det}) shows more clearly the efficiency of learning tracker performance 
online. In almost all sequences the HMMTxD is able to identify and learn which trackers works 
correctly and achieve the performance of at least the best tracker or 
higher (e.g. \textit{motocross1, skating1(low), Volkswagen, singer1, pedestrian3, surfer}). 
Most notable failure cases are caused by the detector failure, e.g. in sequences
\textit{singer2, woman, skating1, basketball, girl\_mov}.

In all other experiments, the abbreviation HMMTxD refers to the combination of all 3
trackers.

\begin{figure*}
         \centering
         \begin{subfigure}[b]{0.32\textwidth}
                 \includegraphics[width=\textwidth]{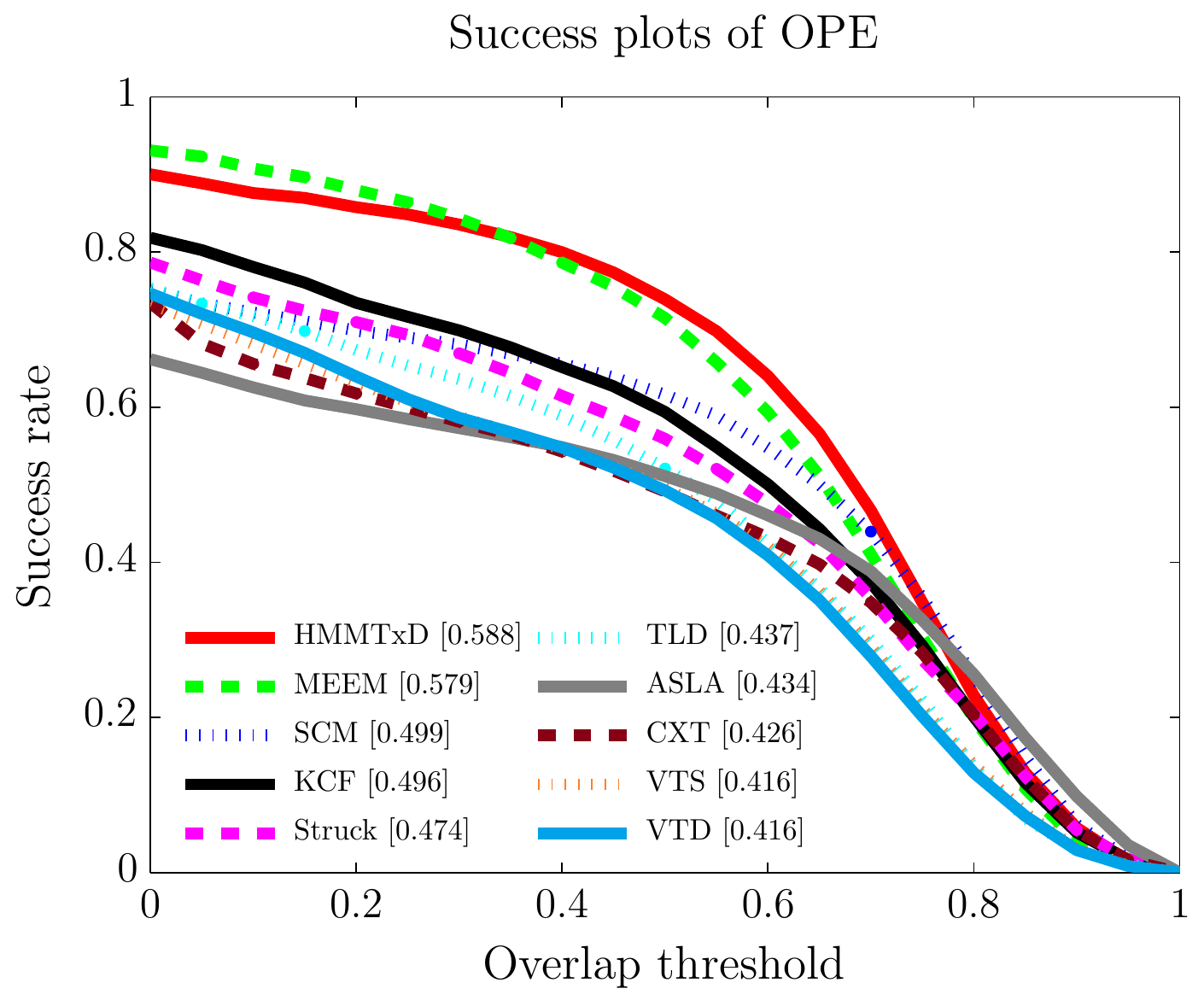}
         \end{subfigure}
         \begin{subfigure}[b]{0.32\textwidth}
                 \includegraphics[width=\textwidth]{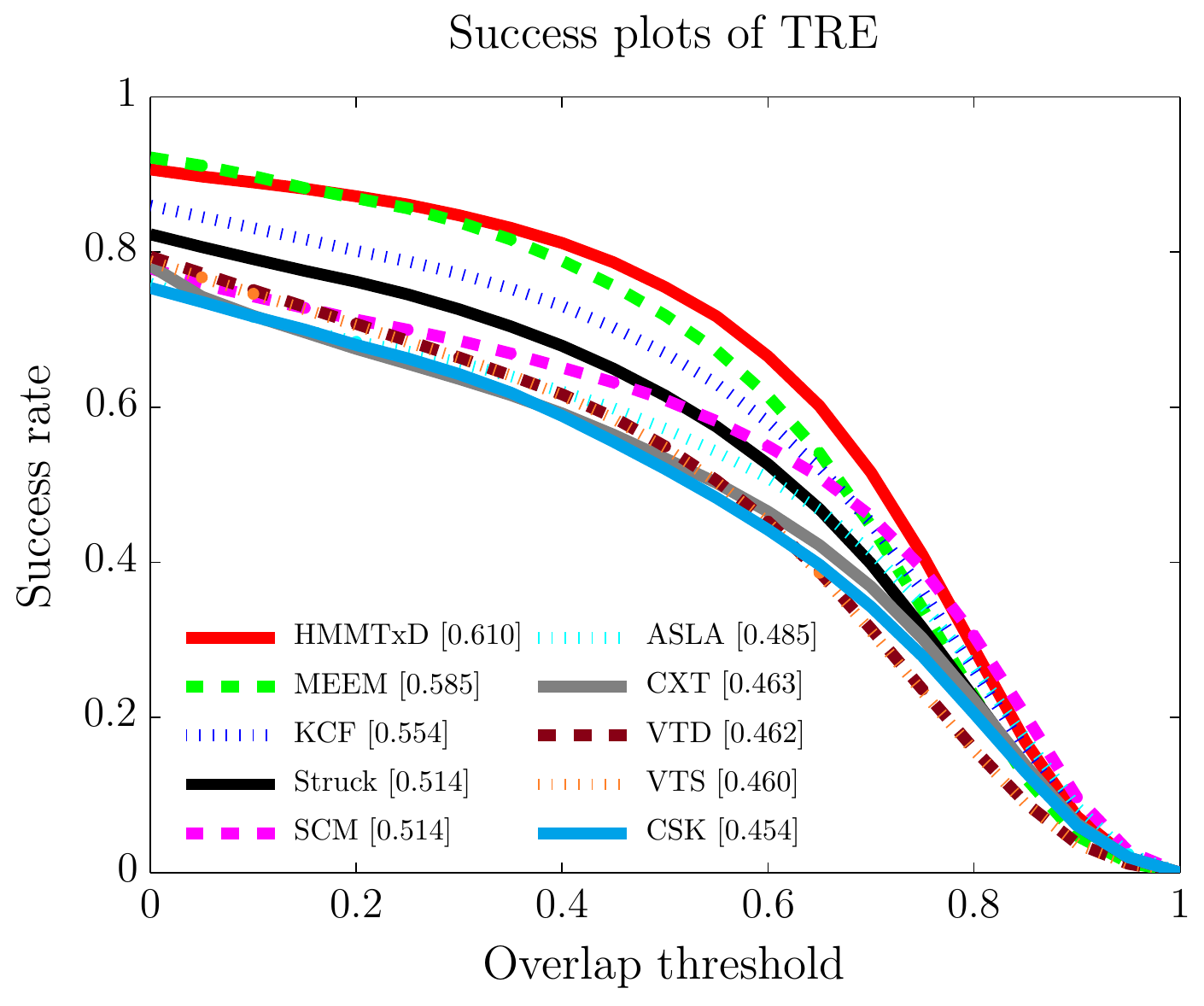}
         \end{subfigure}
         \begin{subfigure}[b]{0.32\textwidth}
                 \includegraphics[width=\textwidth]{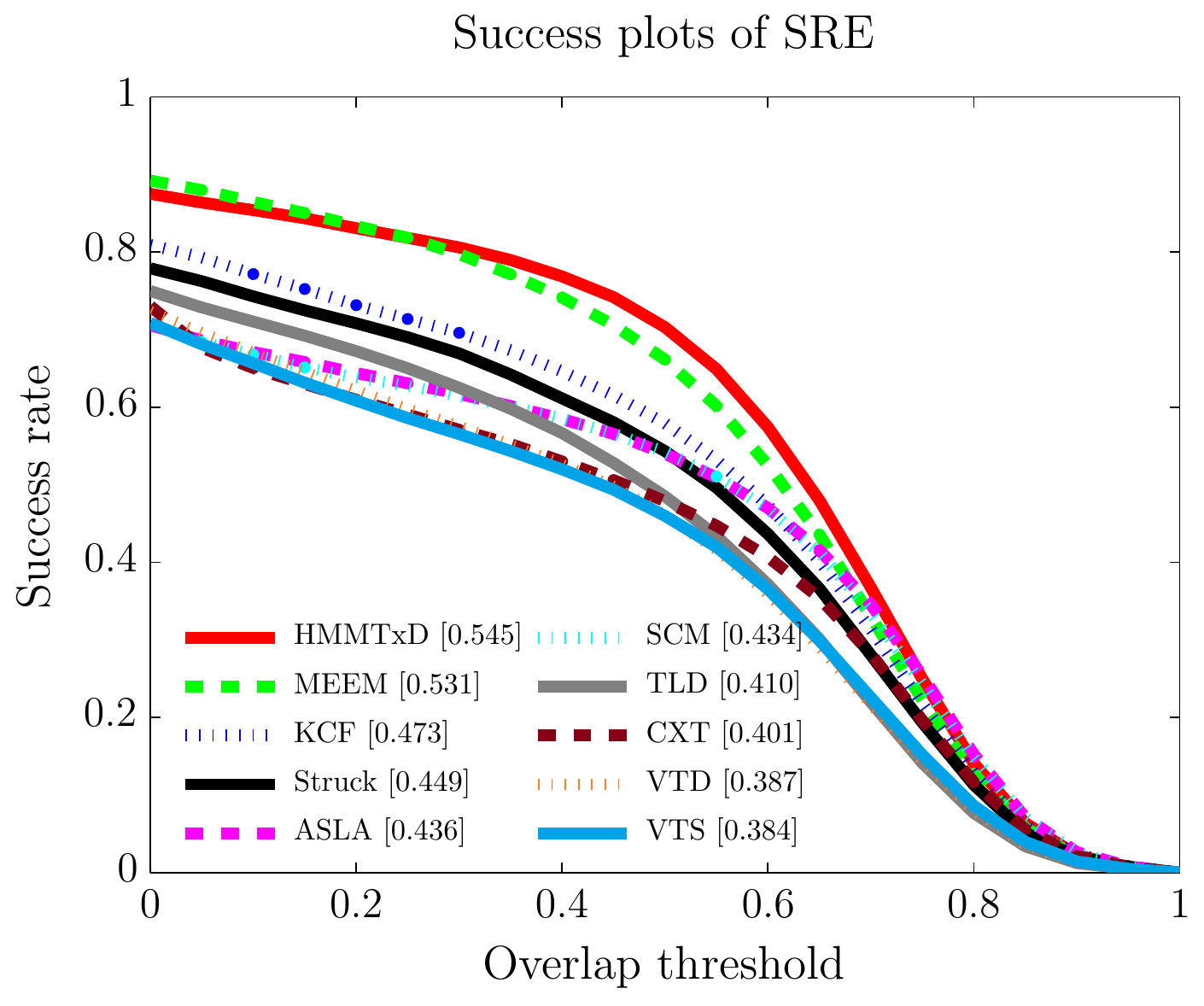}
         \end{subfigure}

         \begin{subfigure}[b]{0.32\textwidth}
                 \includegraphics[width=\textwidth]{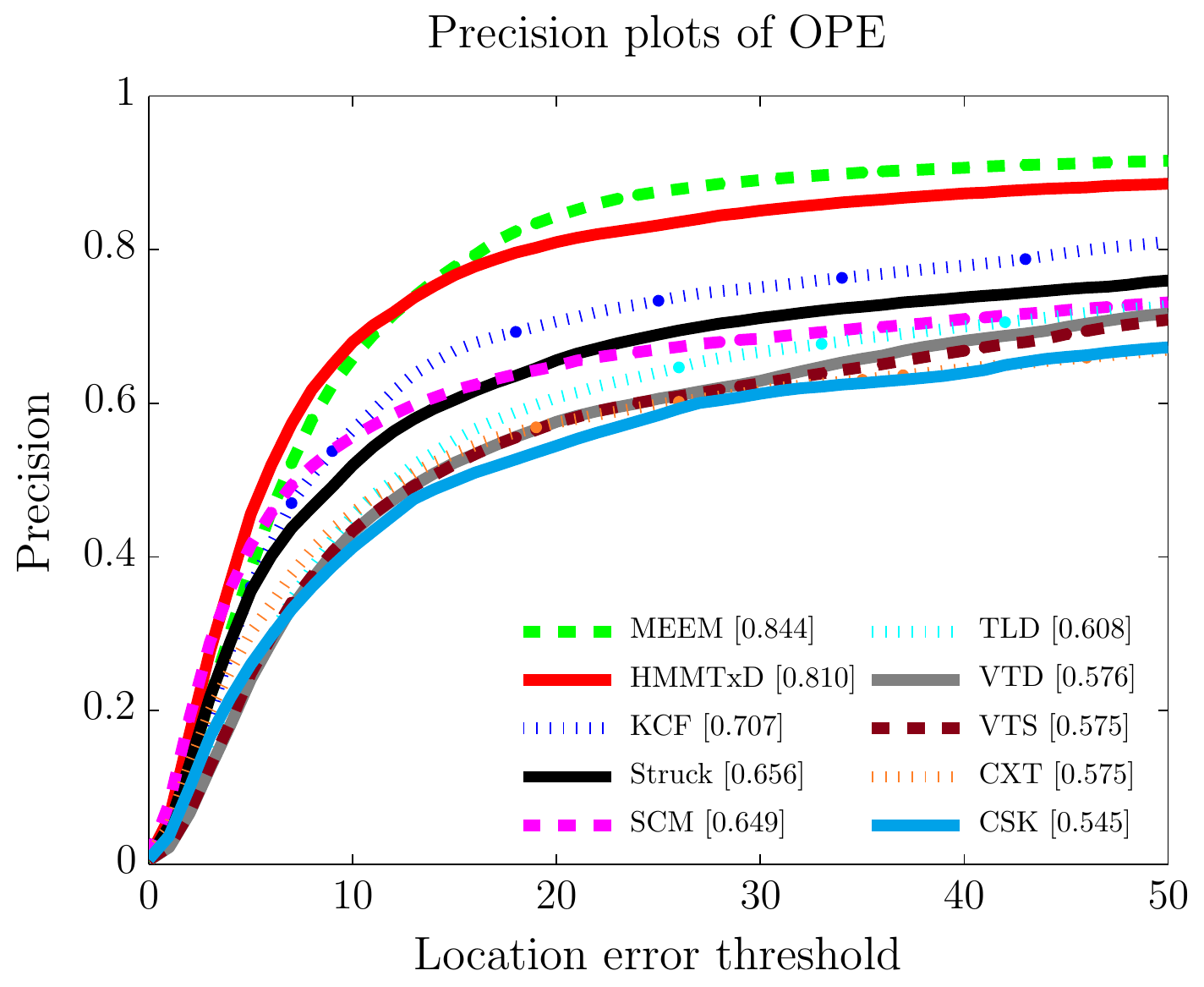}
         \end{subfigure}
         \begin{subfigure}[b]{0.32\textwidth}
                 \includegraphics[width=\textwidth]{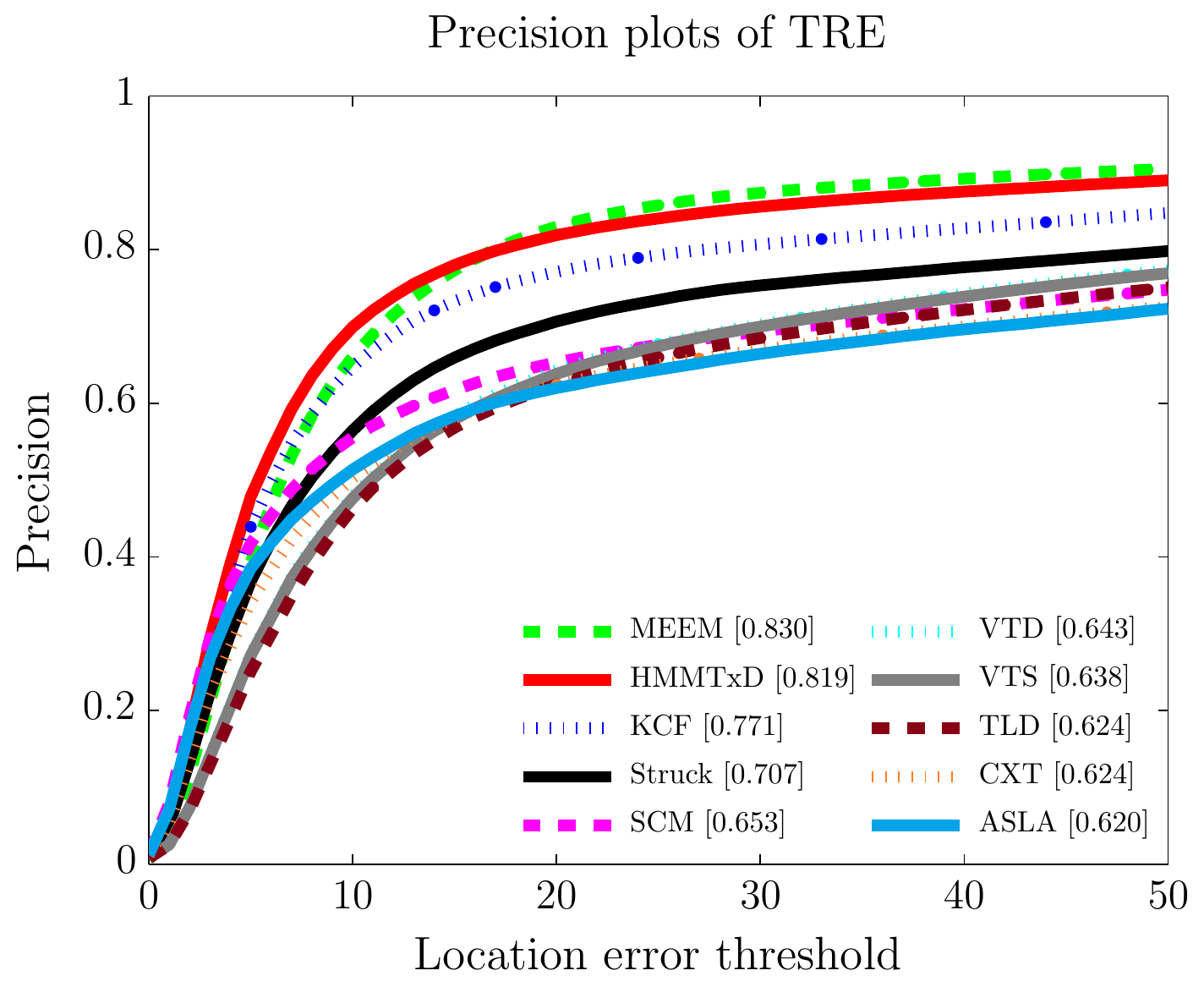}
         \end{subfigure}
         \begin{subfigure}[b]{0.32\textwidth}
                 \includegraphics[width=\textwidth]{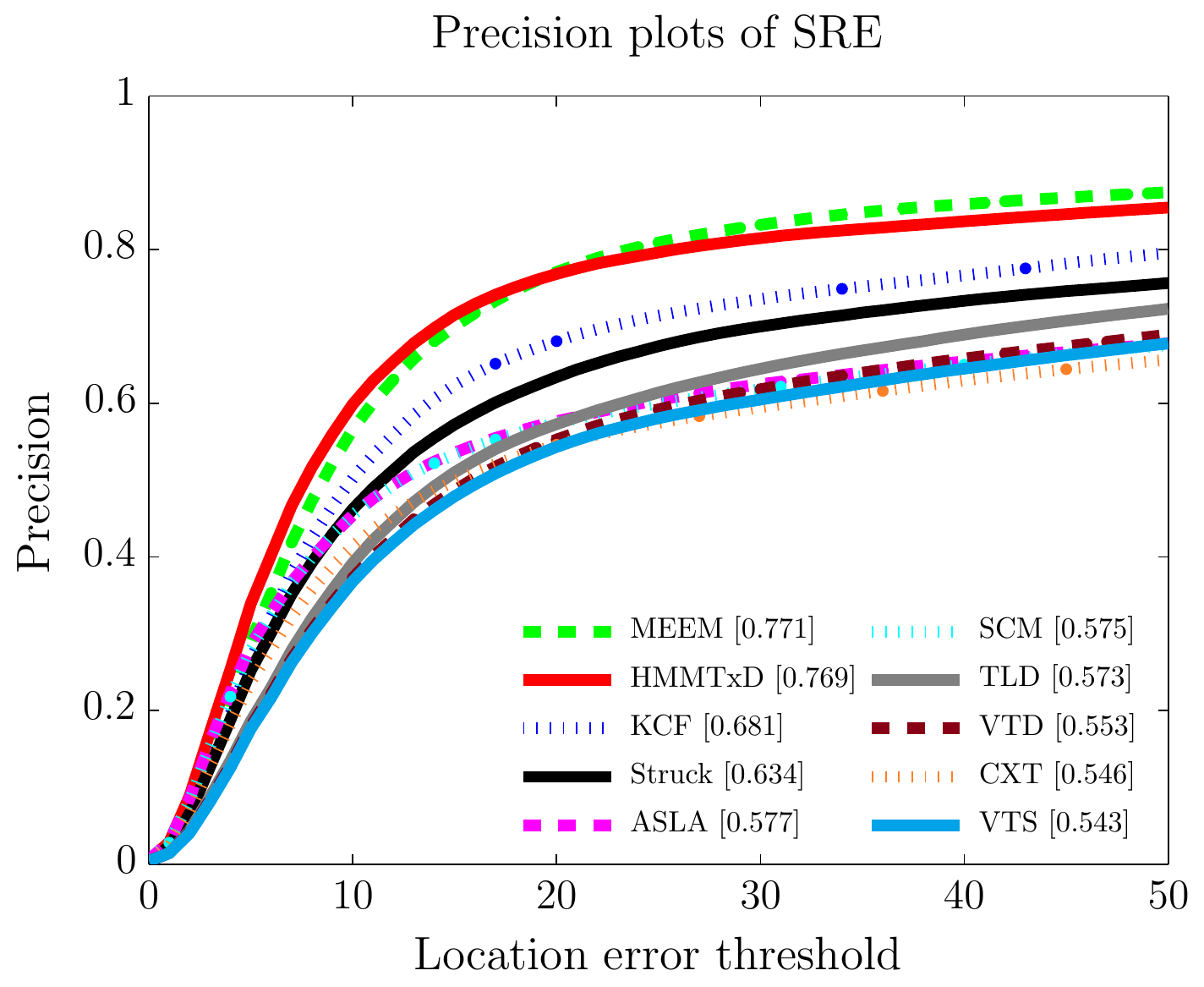}
         \end{subfigure}
\caption{
Evaluation of HMMTxD on the  CVPR2013 Benchmark~\cite{Wu2013}. The top row
shows the success rate as a function of the overlap threshold. The bottom row
shows the precision as a function of the localization error threshold. The
number in the legend is AUC, the area under ROC-curve, which summarizes the
overall performance of the tracker for each experiment.}

\label{fig:cvpr2013Bench}
\end{figure*}

{\bf Evaluation on the CVPR2013 Benchmark}~\cite{Wu2013} 
that contains 50 video sequences. Results on the benchmark have been published
for about 30 trackers. The benchmark defines three types of experiments: (i)
one-pass evaluation (OPE) -- a tracker initialized in the first frame is run
to the end of the sequence, (ii) temporal robustness evaluation (TRE) -- the
tracker is initialized and starts at a random frame, and  (iii) spatial
robustness evaluation (SRE) -- the initialization is perturbed spatially.
Performance is measured by precision (spatial accuracy, i.e. center distance
of ground truth and reported bounding box) and success rate (the number of
frames where overlap with the ground truth was higher than a threshold). The
results are visualized in Fig.~\ref{fig:cvpr2013Bench} where only results of
the 10 top performing trackers are plotted. Together with the tracker from this benchmark, we also added the MEEM~\cite{zhang2014} tracker, which is a recent
state-of-the-art tracker. The proposed HMMTxD outperforms
all trackers in the success rate in all three experiments. Its precision is
comparable to MEEM~\cite{zhang2014} the top performing tracker in terms of precision.
HMMTxD outperforms significantly the OPE results reported in Wang et
al.~\cite{wangg14}, where 5 top performing trackers from this particular
benchmark were used for combination (other experiments were not reported in
the paper).

{\bf VOT2013 benchmark}~\cite{Kristan2013} evaluates trackers on a collection
containing 16  sequences carefully selected from a large pool by a semi-
automatic clustering method. For comparison, results of 27 tracking methods are available and the added MEEM tracker was evaluated by us using default setting from the publicly available source code. The performance is measured by accuracy, average overlap with
the ground truth, and robustness, the number of re-initialization of the
tracker so that it is able to track the whole sequence. Average rank of
trackers is used as an overall performance indicator.

\begin{figure}[!ht]
    \centering
         \begin{subfigure}[b]{0.49\textwidth}
                 \includegraphics[width=\textwidth]{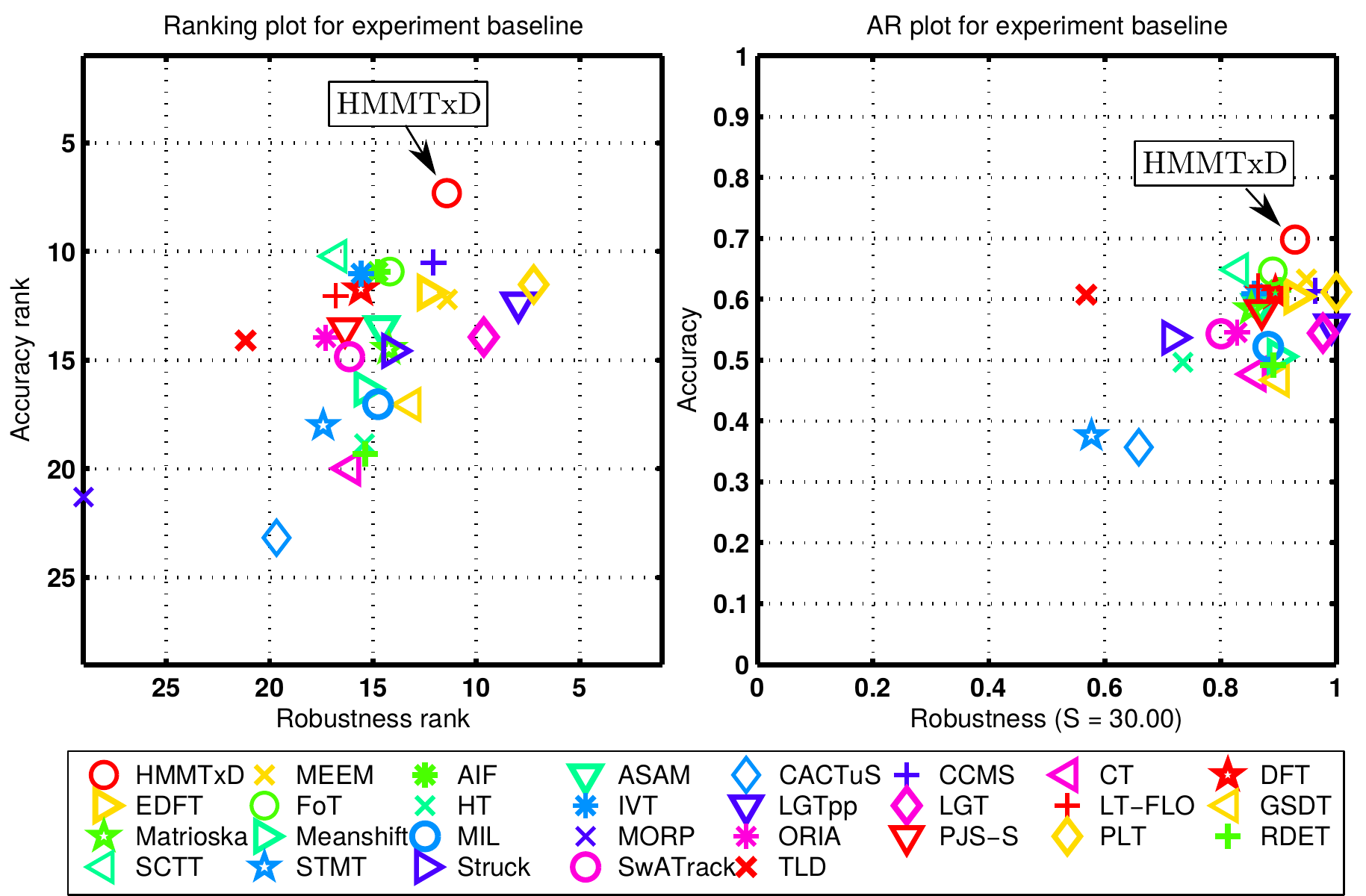}
                 \caption{baseline experiment}
         \end{subfigure}
         \begin{subfigure}[b]{0.49\textwidth}
                 \includegraphics[width=\textwidth]{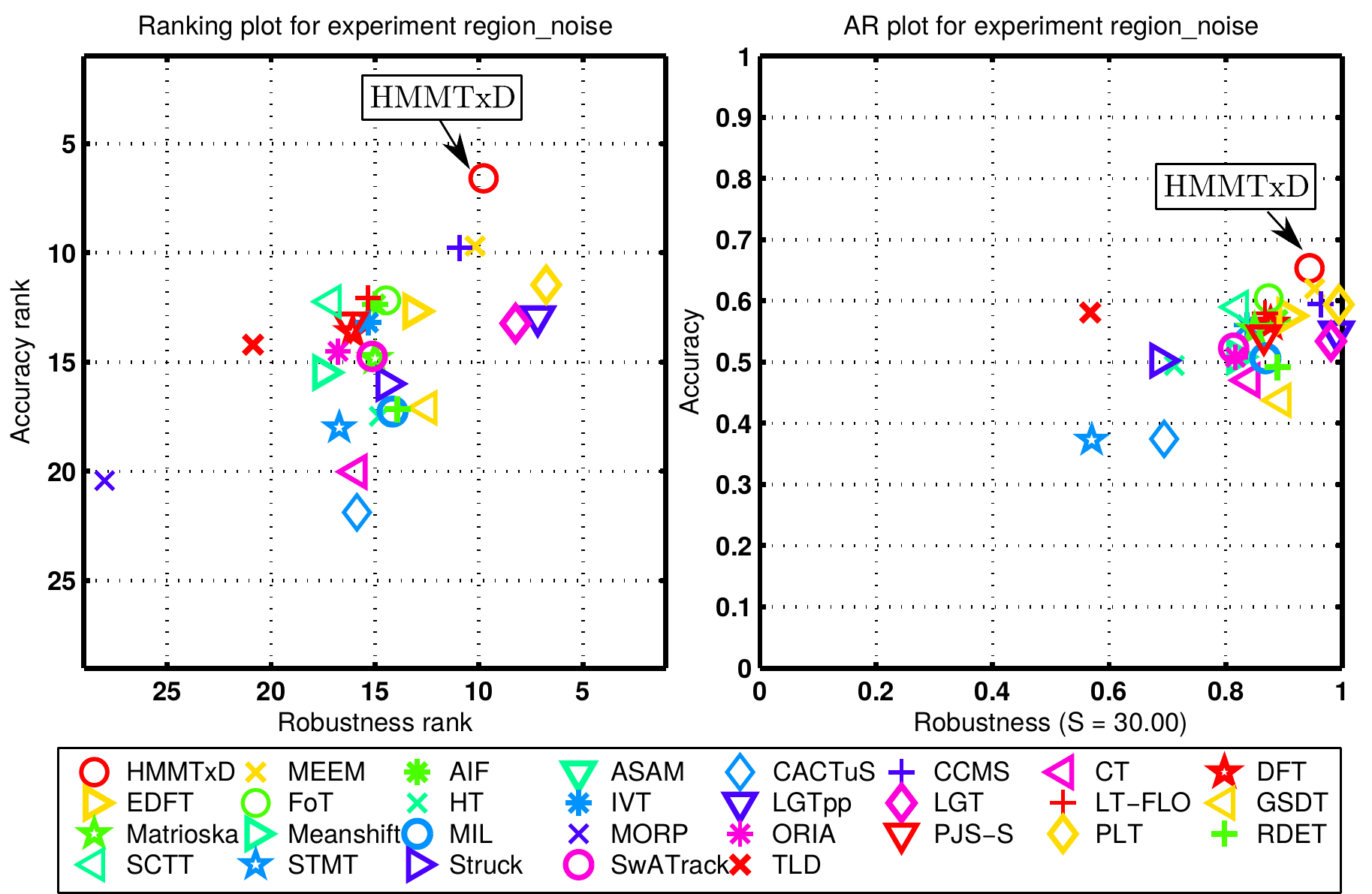}
                 \caption{region-noise experiment}
         \end{subfigure}

\caption{
Evaluation of HMMTxD on the VOT 2013 Benchmark~\cite{Kristan2013}. HMMTxD
result is shown as the red circle. The left plot shows the ranking in accuracy
(vertical axis) and robustness (horizontal axis) and the right plot shows the
raw average values of accuracy and robustness (normalized to the $(0,1)$
interval). For both plots the top right corner is the best performance.}

\label{fig:vot2013Bench}
\end{figure}

In this benchmark, the proposed HMMTxD achieves clearly the best accuracy
(Fig.~\ref{fig:vot2013Bench}). With less than one re-initialization per
sequence it performs slightly worse in terms of robustness due to two reasons.

Firstly, the HMM recognizes a tracker problem with a delay and switching to
other tracker (here even one frame where the overlap with ground truth is zero
leads to penalization) and secondly the VOT evaluation protocol, which require
re-initialization after failure and to forget all previously learned models
(the VOT2013 refer to this as causal tracking), therefore the learned
performance of the trackers is forgotten and has to be learned from scratch.

The results for the baseline and region-noise experiments are shown in
Fig.~\ref{fig:vot2013Bench}. Note that the ranking of the methods differs from
the original publication since two new methods (HMMTxD and MEEM) were added and
the relative ranking of the methods changed.  The top three performing
trackers and their average ranks are HMMTxD ($8.77$), PLT ($9.24$),
LGTpp~\cite{Xiao2013} ($10.11$). MEEM tracker ends up at the fifth place with
average rank $10.87$. The rankings were obtained by the toolkit provided by
the VOT in default settings for baseline and region noise experiments.

\begin{figure*}
    \centering
    \includegraphics[width=\textwidth]{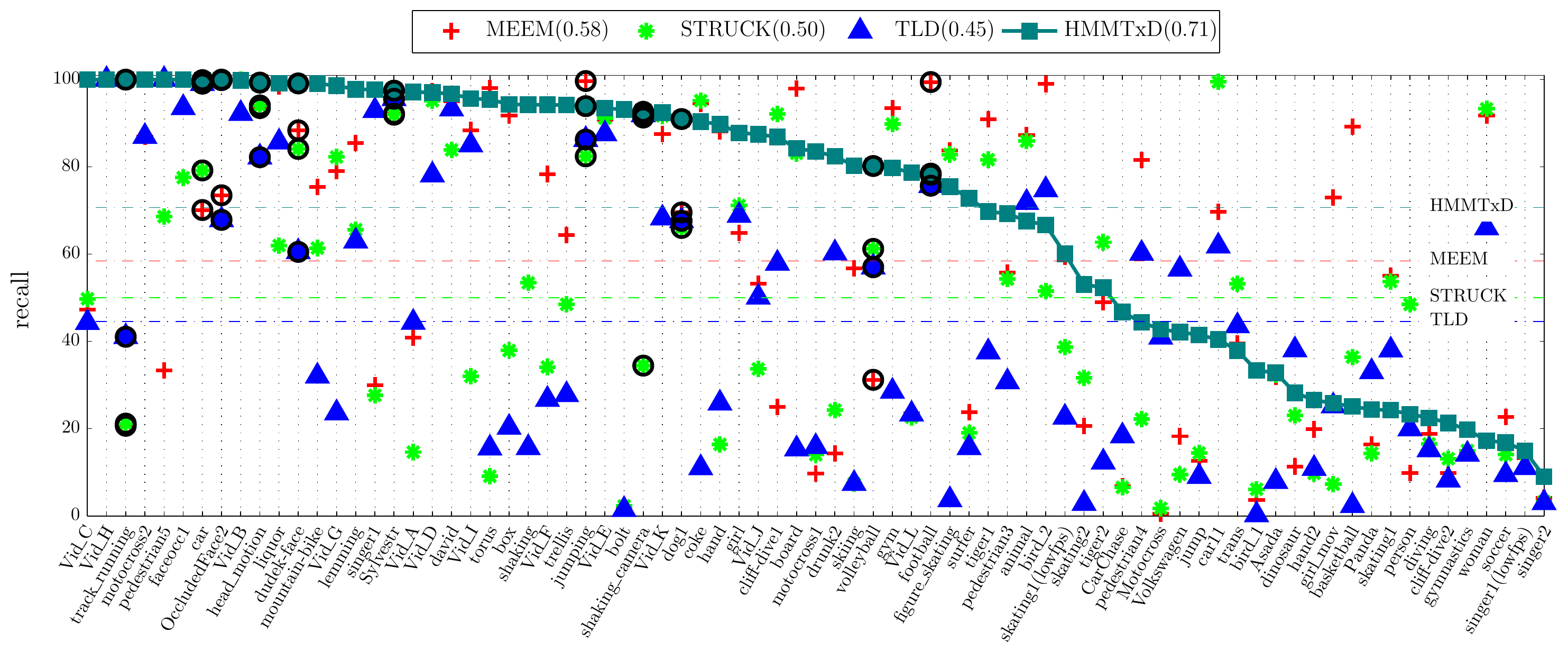}
\caption{
Evaluation of state-of-the-art trackers on the TV77 dataset in terms of
recall, i.e. number of correctly tracked frames. The average recall is shown
by the dashed lines (precise number is in the legend). Black circles mark grayscale sequences. The sequences are ordered by HMMTxD performance.}

\label{fig:tv77Bench}
\end{figure*}

\begin{figure*}[!t]
    \centering
    \includegraphics[width=\textwidth]{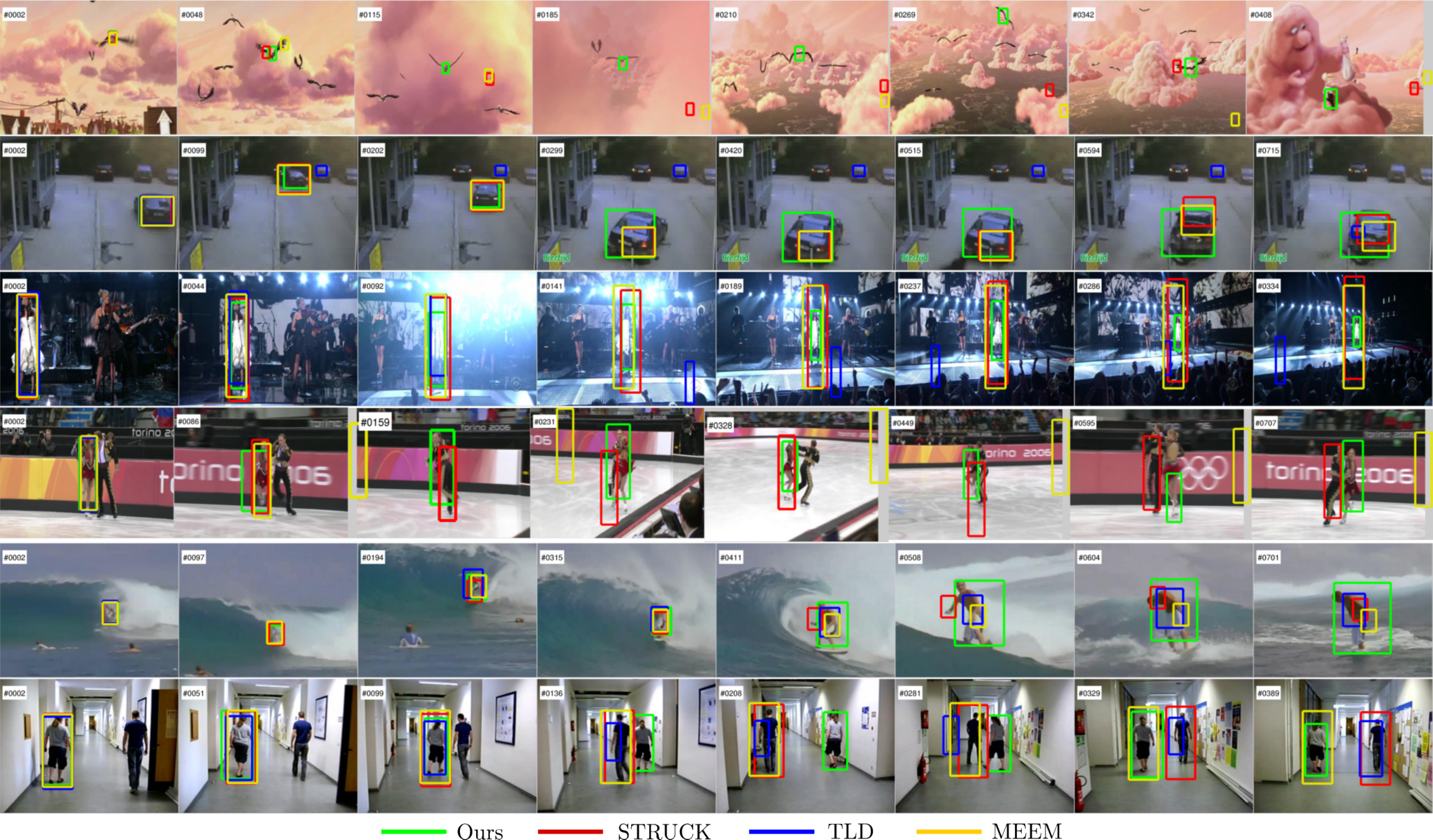}
\caption{
Qualitative comparison of the state-of-the-art trackers on challenging sequences from the 
TV77 dataset (from top \textit{bird\_1, drunk2, singer1, skating2, surfer, Vid\_J}). }
\label{fig:qualitative}
\end{figure*}

The second best performing method on the VOT2013 is the unpublished PLT for
which just a short description is available in \cite{Kristan2013}. PLT is a
variation of structural SVM that uses multiple features (color, gradients).
STRUCK~\cite{Hare2011} and MEEM~\cite{zhang2014} are similar method to the PLT
based on SVM classification. We compared these method with HMMTxD on the
diverse 77 videos along with the TLD~\cite{Kalal2012} which has a similar
design as HMMTxD. HMMTxD outperforms all these methods by a large margin on
average recall -- measured as number of frames where the tracker overlap with
ground truth is higher than $0.5$ averaged over all sequences. Results are
shown in Fig.~\ref{fig:tv77Bench}. Qualitative comparison of these state-of-the-art methods
is shown in Fig.~\ref{fig:qualitative}. Even for sequences with lower recall 
(e.g. \textit{bird\_1, skating2}), the HMMTxD is able to follow the object of interest.

\section{Conclusions}

A novel method called HMMTxD for fusion of multiple trackers has been
proposed.  The method utilizes an on-line trained HMM to estimate the states
of the individual trackers and to fuse a different types of observables provided by the trackers. 
The HMMTxD outperforms its constituent parts (FoT, ASMS, KCF, Detector and its combinations) 
by a large margin and shows the efficiency of the HMM with combination of three trackers.

HMMTxD outperforms all methods included in the CVPR2013 benchmark 
and perform favorably against most recent state-of-the-art tracker. The HMMTxD
also outperforms all method of the VOT2013 benchmark in accuracy, while
maintaining very good robustness, and ranking in the first place in overall
ranking. Experiments conducted on a diverse dataset TV77 show that the HMMTxD
outperforms state-of-the-art MEEM, STRUCK and TLD methods, which are similar
in design, by a large margin. The processing speed of the HMMTxD is $5-10$
frames per second on average, which is comparable with other complex tracking
methods.

\section*{Acknowledgements}
The research was supported by the Czech Science Foundation Project GACR
P103/12/G084 and by the Technology Agency of the Czech Republic project TE01020415 (V3C -- Visual Computing Competence Center).

\appendix
\section{Forward-Backward Procedure for Modified Baum-Welch Algorithm}
\label{app:fb}

Let us assume the HMM with $N$ possible states $\{s_1,s_2,\ldots,s_N\}$, the
matrix of state transition probabilities $A=\{a_{ij}\}_{i,j=1}^{N}$, the
vector of initial state probabilities $\pi=(1,0,0,\ldots,0)$, the initial
state $s_1=(1,1,\ldots,1)$, a sequence of observations
$\mathbb{X}=\{X_t\}_{t=1}^{T},X_t\in R^{m}$ and $F=\{f_i(x)\}_{i=1}^{N}$ the
system of conditional probability densities of observations conditioned on
$S_t=s_i$.

Let  $0=t_0<t_1<t_2\ldots <t_K\leq T$ be a sequence of detection times,
$\mathbb{S}=\{S_{t_{k}}=s_{i_k},\{t_k\}_{k=1}^{K}\}$ be observed states of
Markov chain, marked by the detector, and $S_{t_{k}+1}=s_1$ for $0\leq k\leq
K$.

The forward variable for the Baum-Welch algorithm is defined as follows.
Let $1\leq i \leq N, 1\leq k \leq K, t_{(k-1)}<t \leq t_k $ and
\begin{equation}
\label{eq:A1}
\alpha_t(i)=P(X_{t_{(k-1)}+1},\ldots , X_t,S_{t}=s_i|\lambda) \text { then}
\end{equation}
 \begin{equation}
\alpha_{t_{(k-1)}+1}(1)=f_1(X_{t_{(k-1)}+1}),
\end{equation}
\begin{equation}
\alpha_{t_{(k-1)}+1}(i)=0 \; \text{for} \; i\neq 1
\end{equation}
and for $t_{(k-1)}<t< t_k$
\begin{equation}
\alpha_{(t+1)}(i)=\sum_{j=1}^{N}\alpha_t(j)a_{ji}f_i(X_{t+1}),
\end{equation}
\begin{equation}
P(S_{t}=s_i|X_{1},\ldots X_{t},S_{t_1},S_{t_2},\ldots, S_{t_{(k-1)}},\lambda)= \frac{\alpha_{t}(i)}{\sum_{j=1}^{N}  \alpha_ {t}(j)}.
\label{eq:app_chain}
\end{equation}
For $t_{K}<t< T$ the forward variable is in principle the same as above  with $t_{(k-1)}=t_K$. So
\begin{equation}
P(X_{t_{K}+1},\ldots , X_{T}|\lambda)=\sum_{i=1}^{N} \alpha_{T}(i)
\end{equation}
\begin{equation}
\label{eq:A7}
P(\mathbb{X},\mathbb{S}|\lambda)= \prod_{k=1}^{K}\alpha_{t_k}(i_k) * \sum_{i=1}^{N}\alpha_{T}(i) \;\;\text{ where } S_{t_k}=s_{i_k}.
\end{equation}
The backward variable for $ t_{(k-1)}<t< t_k$ is
\begin{equation}
\beta_t(i)=P(X_{t+1},\ldots ,X_{t_k},S_{t_k}|S_{t}=s_i,\lambda),
\end{equation}
where $\beta_{t_k}(i_k)=1$ and $\beta_{t_k}(i)=0$ for $i\neq i_k$
and
\begin{equation}
\beta_{t}(i)=\sum_{j=1}^{N}a_{ij} f_j(X_{t+1})\beta_{t+1}(j) .
\end{equation}
For $t_{K}<t< T$ the backward variable is in principle the same as above where
$\beta_{T}(i)=1$ for $1\leq i \leq N$.

Given the forward and backward variables, we get the following probabilities,
that are used to update parameters of HMM. For $0<t<T \text{ and } t\neq
t_k,1\leq k \leq K$

\begin{equation}
P(S_{t}=s_i,S_{t+1}=s_j|\mathbb{X},\mathbb{S},\lambda)=
\end{equation}
\begin{equation}
\frac{\alpha_{t}(i)a_{ij} f_j(X_{t+1})\beta_{(t+1)}(j)}
{\sum_{k=1}^{N}\sum_{l=1}^{N}\alpha_{t}(k)a_{kl} f_l(X_{t+1})\beta_{t+1}(l)}
\end{equation}

and for $0<t\leq T$
\begin{equation}
P(S_{t}=s_i|\mathbb{X},\mathbb{S},\lambda)=\frac{\alpha_{t}(i)\beta_{t}(i)}
{\sum_{j=1}^{N}\alpha_{t}(j)\beta_{t}(j)}.
\end{equation}

The final equation for the update of transition probabilities $A$ of HMM is as follows.
\begin{equation}
\hat{a}_{ij}=\frac{\text{expected number of transitions from state $s_i$ to state $s_j$}}{\text{expected number of transitions from state $s_i$}}
\end{equation}
\begin{equation}
=\frac{\sum_{(t=1\text{ and } t\neq t_{k},1\leq k \leq K) }^{T-1}P(S_{t}=s_i,S_{t+1}=s_j|\mathbb{X},\mathbb{S},\lambda)}
{\sum_{(t=1\text{ and } t\neq t_{k},1\leq k \leq K)}^{T-1}P(S_{t}=s_i|\mathbb{X},\mathbb{S},\lambda)}.
\end{equation}

\section{Generalized Method of Moments}
\label{app:moments}

For a simplification let us assume HMM with one-dimensional observed random
variables $\{X_t\}_{t=1}^{+\infty}, X_t \in R$. The sequence
$\{X_t-E(X_t|X_1,X_2,\ldots,X_{t-1})\}_{t=1}^{+\infty}$ is a martingale
difference series where

\begin{equation}
E(X_t|X_1,X_2,\ldots,X_{t-1})=\sum_{i=1}^{N}E(X_t|X_1,X_2,\ldots,X_{t-1},S_t=i)P(S_t=i)
\end{equation}
\begin{equation}
=\sum_{i=1}^{N}E(X_t|S_t=i)P(S_t=i).
\end{equation}

Under the assumption that $\{X_t\}_{t=1}^{+\infty}$ are uniformly bounded
random variables i.e. $|X_t|<c, c \in (0,+\infty)$ for all $t\geq1$, the
strong law of large numbers for a sum of martingale differences can be
used(see Theorem 2.19 in ~\cite{hall1980}). So

\begin{equation}
\lim_{T \to  +\infty} \frac{1}{T}\sum_{t=1}^{T}[X_t-\sum_{i=1}^{N}E(X_t|S_t=i)P(S_t=i)]= 0 \text{  almost surely}.
\end{equation}
Let us denote $\mu_i=E(X_t|S_t=i)$  for  $1 \leq t \leq T$
and $\hat{\mu}_i$ the estimate of $\mu_i$ based on the modified method of moments.
The estimate $\hat{\mu}_i$ is a solution of a following equation w.r.t. $\mu_i$

\begin{equation}
\frac{1}{T}\sum_{t=1}^{T}X_t =\frac{1}{T}\sum_{t=1}^{T}\sum_{i=1}^{N}\mu_i P(S_t=i).
\label{eq:app1}
\end{equation}

Having one equation for $N$ unknown variables $\mu_i,1\leq i\leq N$ it is
necessary to add some constrains to get a unique solution. We propose to
minimize

\begin{equation}
\sum_{t=1}^{T}\sum_{i=1}^{N}(X_t-\mu_i)^2 P(S_t=i),
\end{equation}
w.r.t. $\mu_i,1\leq i\leq N$ giving
\begin{equation}
\hat{\mu}_i=\frac{\sum_{t=1}^{T} X_t P(S_{t}=s_i)}
{\sum_{t=1}^{T}P(S_{t}=s_i)}
\end{equation}
which satisfy the moment equation (\ref{eq:app1}).
The same way of reasoning can be used for higher moments of $\{X_t\}_{t=1}^{T}$.
For example using $\{(X_t)^2\}_{t=1}^{T}$ we get estimates $\hat{\sigma}_i^2$ for
$\sigma_i^2=var(X_t|S_t=i)$ for $1 \leq t \leq T$,

\begin{equation}
\hat{\sigma}_i^2=\frac{\sum_{t=1}^{T} (X_t-\hat{\mu}_i)^2 P(S_{t}=s_i)}
{\sum_{t=1}^{T}P(S_{t}=s_i)}.
\end{equation}

In the HMMTxD $m$-dimensional observed random variables
$X_t=(X_t^1,X_t^2,\ldots, X_t^m)$ are assumed, each of them having beta-
distribution and being conditionally independent. There are well-known
relations for a mean value $EX$ and a variance $varX$ of a random variable $X$
having beta distribution and its shape parameters $(p,q)$

\begin{equation}
p=EX \left( \frac{EX(1-EX)}{varX}-1\right)
\end{equation}
and
\begin{equation}
q=(1-EX) \left( \frac{EX(1-EX)}{varX}-1\right).
\end{equation}

Using the modified method of moments gives
\begin{equation}
\hat{\mu}_i^j=\frac{\sum_{t=1}^{T}X_t^j P(S_{t}=s_i|\mathbb{X},\mathbb{S},\lambda)}
{\sum_{t=1}^{T}P(S_{t}=s_i|\mathbb{X},\mathbb{S},\lambda)}
\end{equation}
and
\begin{equation}
(\hat{\sigma}_i^j)^2=\frac{\sum_{t=1}^{T}
(X_t^j-\hat{\mu}_i^j)^2P(S_{t}=s_i|\mathbb{X},\mathbb{S},\lambda)}
{\sum_{t=1}^{T}P(S_{t}=s_i|\mathbb{X},\mathbb{S},\lambda)}.
\end{equation}
Then
\begin{equation}
\hat{p}_i^j={\hat{\mu}_i^j}\left( \frac{\hat{\mu}_i^j(1-\hat{\mu}_i^j)}{(\hat{\sigma}_i^j)^2}-1\right)
\end{equation}
and
\begin{equation}
\hat{q}_i^j=(1-\hat{\mu}_i^j) \left( \frac{\hat{\mu}_i^j(1-\hat{\mu}_i^j)}{(\hat{\sigma}_i^j)^2}-1\right).
\end{equation}

If we assume in our model $\lambda=(A,F)$ that for some $\{(i_r,j_r) \in
\{1,2,\ldots,N\} \times \{1,2,\ldots,m\}: p_{i_r}^{j_r}=p,q_{i_r}^{j_r}=q
\}_{r=1}^R$ then

\begin{equation}
\hat{p}=\hat{\mu} \left( \frac{\hat{\mu}(1-\hat{\mu})}{\hat{\sigma}^2}-1\right)
\end{equation}
and
\begin{equation}
\hat{q}=(1-\hat{\mu}) \left( \frac{\hat{\mu}(1-\hat{\mu})}{\hat{\sigma}^2}-1\right)
\end{equation}
where
\begin{equation}
\hat{\mu}=\frac{\sum_{r=1}^R\sum_{t=1}^{T} X_t^{j_r} P(S_{t}=s_{i_r}|\mathbb{X},\mathbb{S},\lambda)}
{\sum_{r=1}^R\sum_{t=1}^{T}P(S_{t}=s_{i_r}|\mathbb{X},\mathbb{S},\lambda)}
\end{equation}
and
\begin{equation}
\hat{\sigma}^2=\frac{\sum_{r=1}^R\sum_{t=1}^{T} (X_t^{j_r}-\hat{\mu})^2    P(S_{t}=s_{i_r}|\mathbb{X},\mathbb{S},\lambda)}
{\sum_{r=1}^R\sum_{t=1}^{T}P(S_{t}=s_{i_r}|\mathbb{X},\mathbb{S},\lambda)}.
\end{equation}




\section*{References}
\small{
\bibliographystyle{elsarticle-num} 
\bibliography{main}
}

\end{document}